\documentclass[runningheads]{llncs}

\usepackage{eccv}

\usepackage{eccvabbrv}

\usepackage{graphicx}
\usepackage{algorithm}
\usepackage{algorithmicx,algpseudocode}
\usepackage{color, colortbl}
\usepackage{booktabs}
\usepackage{enumitem}
\usepackage{xcolor}
\usepackage{amssymb}
\usepackage{amsmath}
\usepackage{wrapfig}
\usepackage[skip=2.5pt]{caption}
\usepackage{arydshln}
\usepackage{multicol}

\definecolor{Gray}{gray}{0.9}

\newcommand\given[1][]{\:#1\vert\:}
\newcommand\norm[1]{\left\lVert#1\right\rVert}

\usepackage[accsupp]{axessibility}  

\usepackage{hyperref}

\usepackage{orcidlink}

\begin{document}

\title{DreamMotion: Space-Time Self-Similar Score Distillation for Zero-Shot Video Editing}

\titlerunning{DreamMotion}

\author{Hyeonho Jeong\inst{1}\orcidlink{0000-0002-6864-4190} \and
Jinho Chang\inst{1}\orcidlink{0000-0002-7426-8304} \and
Geon Yeong Park\inst{2}\orcidlink{0009-0006-7522-4553} \and
Jong Chul Ye\inst{1,2}\orcidlink{0000-0001-9763-9609}}

\authorrunning{Jeong et al.}

\institute{Kim Jaechul Graduate School of AI, KAIST, South Korea \and
Dept. of Bio and Brain Engineering, KAIST, South Korea \\
\email{\{hyeonho.jeong, jinhojsk515, pky3436, jong.ye\}@kaist.ac.kr}\\
Project page: \url{https://hyeonho99.github.io/dreammotion} \\
}

\maketitle
\begin{center}
    \centering
    \captionsetup{type=figure}
    \includegraphics[width=\textwidth]
    {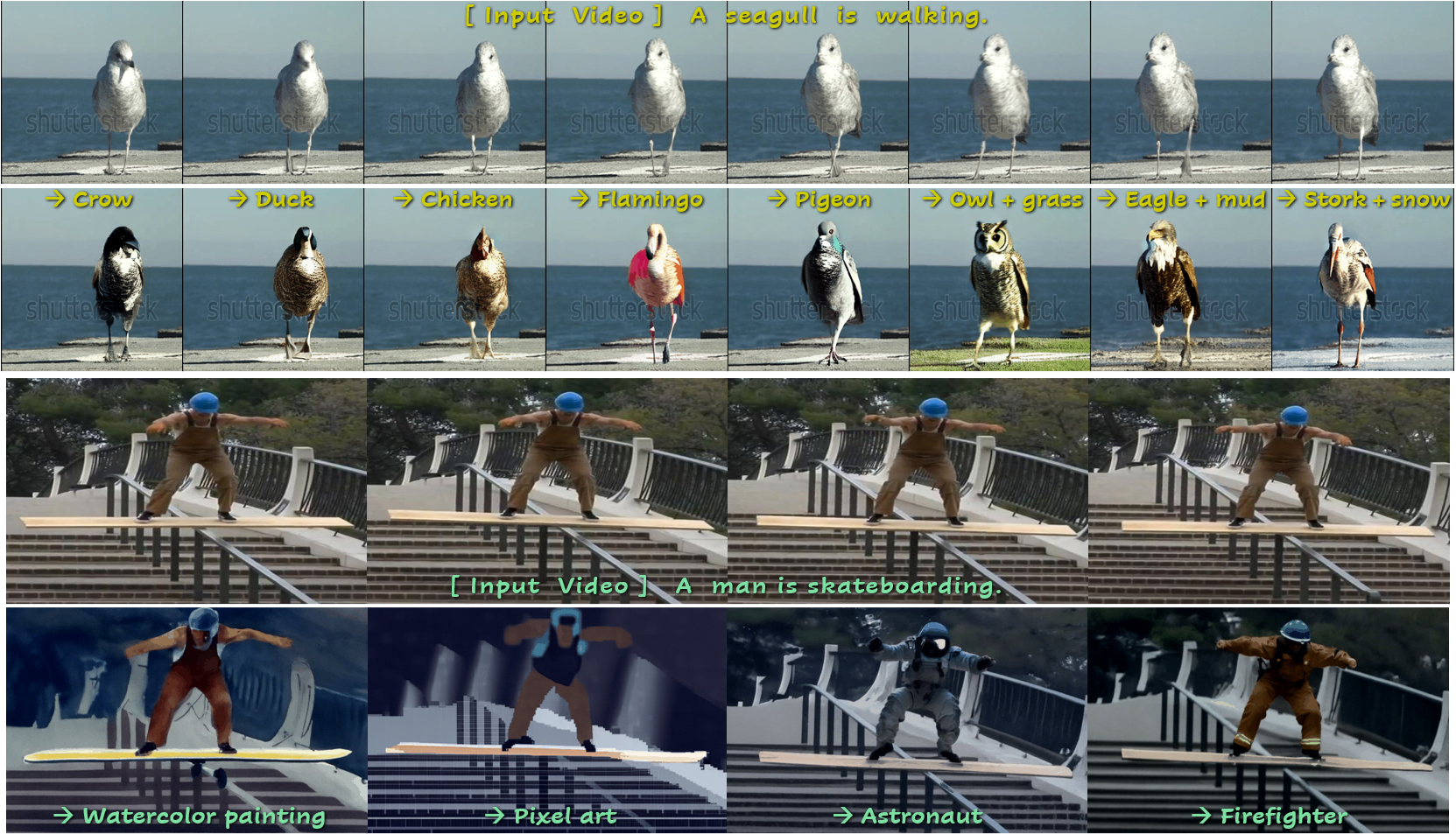}
    \captionof{figure}{
    \textbf{Zero-shot video editing results.}
    The second row presents videos produced with our method with a non-cascaded video diffusion model, while those in the bottom row are from a cascaded model. 
    For a full display of results, visit our \textit{\href{https://hyeonho99.github.io/dreammotion}{project page}}.
    }
    \label{figure: teaser}
\end{center}

\begin{abstract}
Text-driven diffusion-based video editing presents a unique challenge not encountered in image editing literature: establishing real-world motion. 
Unlike existing video editing approaches,
here we focus on score distillation sampling to circumvent the standard reverse diffusion process and initiate optimization from videos that already exhibit natural motion.
Our analysis reveals that while video score distillation can effectively introduce new content indicated by target text, it can also cause significant structure and motion deviation.
To counteract this, we propose to match the space-time self-similarities of the original video and the edited video during the score distillation.
Thanks to the use of score distillation, our approach is model-agnostic, which can be applied for both cascaded and non-cascaded video diffusion frameworks.
Through extensive comparisons with leading methods, our approach demonstrates its superiority in altering appearances while accurately preserving the original structure and motion.
\end{abstract}

\section{Introduction}
\label{sec:intro}
Building upon the progress in diffusion models \cite{sohl2015deep, ho2020denoising, song2020score}, the advent of large-scale text-image pairs \cite{schuhmann2022laion} brought an unprecedented breakthrough in text-driven image generative tasks.
In particular, real-world image editing has undergone significant evolution, supported by foundational Text-to-Image (T2I) diffusion models \cite{rombach2022high, ramesh2022hierarchical,saharia2022photorealistic, podell2023sdxl}.
However, extending the success of diffusion-based image editing to video editing introduces a significant challenge: modeling temporally consistent, real-world motion throughout the reverse diffusion process.

Existing methods leveraging T2I diffusion models typically start by inflating attention layers to attend to multiple frames simultaneously \cite{wu2023tune, khachatryan2023text2video, wang2023zero, qi2023fatezero, chen2023control, wu2023lamp, zhang2023controlvideo, cong2023flatten, jeongground}. 
Yet, this technique falls short of achieving smooth and complete motion, as it depends on the implicit preservation of motion through the inflated attention layers.
As a result, a commonly adopted solution is to employ additional visual hints that explicitly guide the reverse diffusion process.
One strategy is to use attention map guidance, for example, by injecting self-attention maps \cite{ceylan2023pix2video, qi2023fatezero} or manipulating cross-attentions \cite{liu2024video}.
Other works attempt to integrate the denoising process with spatially-aligned structural cues, like depth or edge maps.
For example, pre-trained adapter networks such as ControlNet \cite{zhang2023adding} or GLIGEN \cite{li2023gligen} have been transferred from image to video domain, achieving structure-consistent outputs \cite{chen2023control, zhang2023controlvideo, hu2023videocontrolnet, jeongground}.

\begin{figure}[!t]
    \centering
    \includegraphics[width=\textwidth]
    {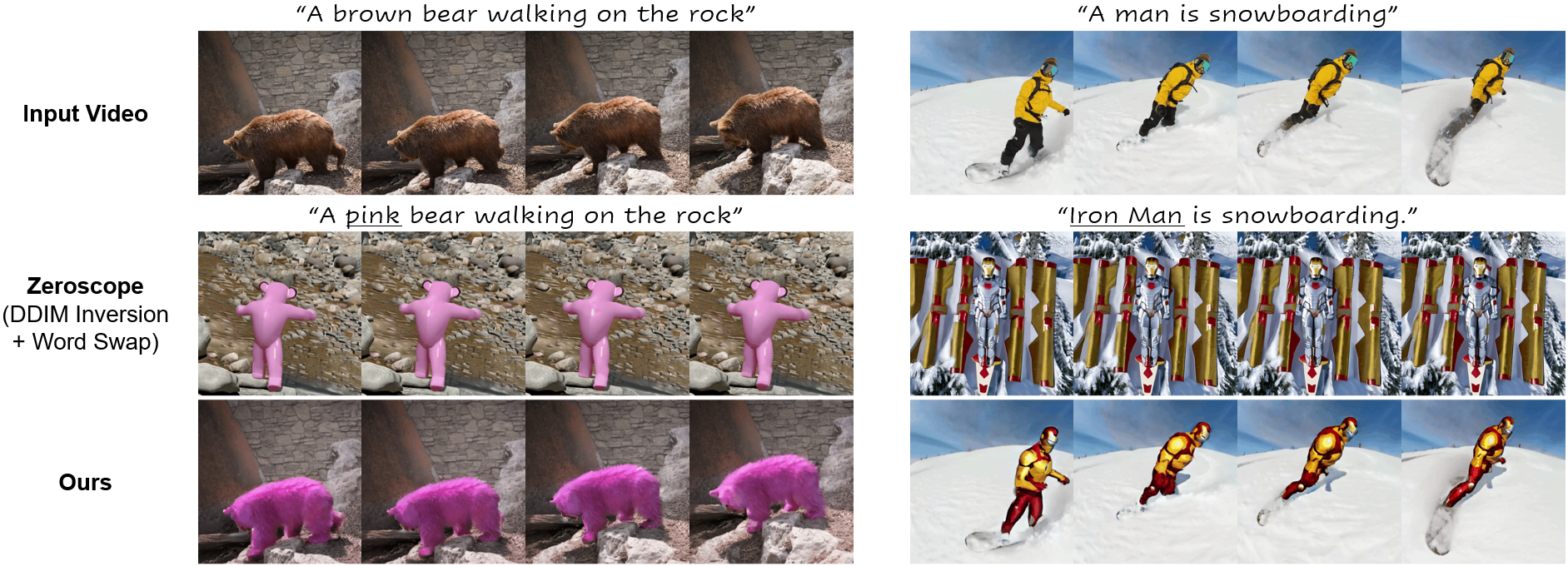}
    \caption{ 
    Ancestral sampling-based zero-shot video editing fails to capture complex, real-world motion in the generated videos.
    }
    \label{figure: ddim}
\end{figure}

Even with the presence of pre-trained Text-to-Video (T2V) diffusion models, zero-shot video editing still poses a significant hurdle since publicly available T2V models \cite{wang2023modelscope, zeroscope} lack sufficiently rich temporal priors to accurately depict real-world motion in the generated videos, as illustrated in Fig. \ref{figure: ddim}.
Thus, recent endeavors often adopt a self-supervised strategy of finetuning pre-trained model weights on the motion presented in an input video \cite{zhao2023motiondirector, jeong2024vmc, park2024spectral, wei2023dreamvideo, zhang2023motioncrafter}. 
Whether employing T2I or T2V models, the conventional reverse diffusion process —beginning with standard Gaussian noise or, at most, inverted latent representations— struggles to reprogram complex, real-world motion, unless supplemented by additional visual conditions or by overfitting the spatial-temporal priors to a particular video.

\begin{figure}[t]
    \centering
    \includegraphics[width=\textwidth]
    {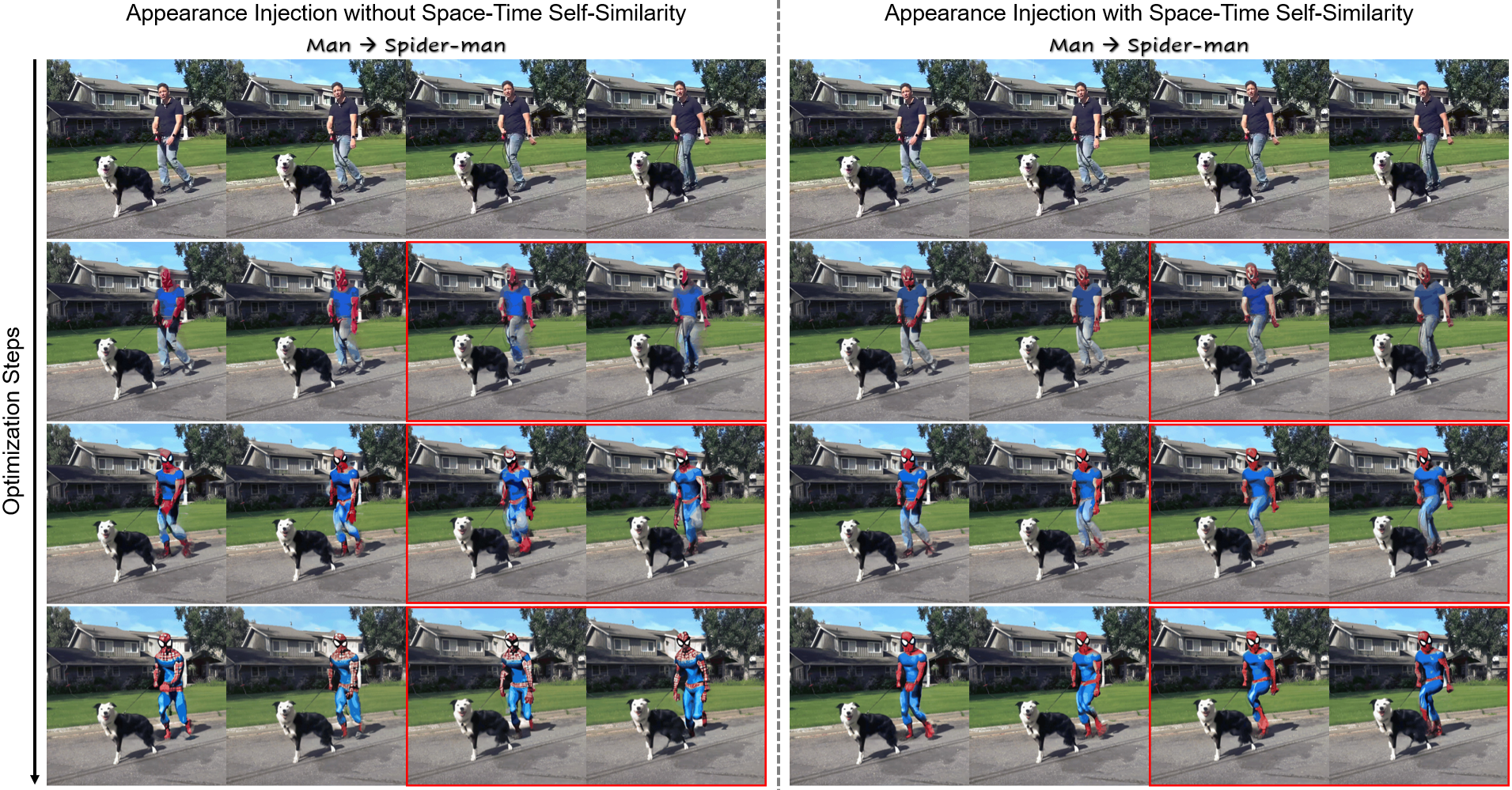}
    \caption{ 
    \textbf{Optimization progress visualization.}
    The proposed self-similarity regularization effectively preserves the structure and motion of the original video.
    }
    \label{figure: progress}
\end{figure}

To this end, we propose to diverge from the previous video editing literature.
Our approach, DreamMotion, deliberately avoids the standard denoising process (ancestral sampling), and instead leverages the Score Distillation Sampling (SDS, \cite{poole2022dreamfusion}) grounded optimization to edit a video.
Specifically, starting from an input video with temporally consistent, natural motion, we attempt to progressively modify the video's appearance while maintaining the integrity of the motion.
In specific, our framework gradually injects target appearance to the video using Delta Denoising Score (DDS, \cite{hertz2023delta}) gradients within T2V diffusion models.
During this procedure, we filter the gradients with additional binary mask conditions to avoid blurriness and over-saturation.
While this optimization effectively infuses the targeted appearance, it tends to accumulate structural errors, resulting in deviations in motion across the final output frames {(see Fig. \ref{figure: progress})}.
To address this, we present self-similarity-based space-time regularization methods.
More specifically, by aligning the spatial self-similarity of diffusion features between the original and edited videos, we preserve structure and motion integrity while seamlessly modifying the appearance.
Furthermore, ensuring temporal self-similarity between the two features facilitates effective temporal smoothing, preventing potential distortions in areas subjected to optimization.
Our methodology is applied to both cascaded and non-cascaded video diffusion models, showcasing its wide applicability across different video editing frameworks.

In summary, DreamMotion offers the following key contributions:
\vspace{-5pt}
\begin{itemize}[noitemsep]
    \item A pioneering zero-shot framework that distills video score from text-to-video diffusion priors to inject target appearance.
    \item A novel space-time regularization that aligns spatial self-similarity to minimize structural deviations and temporal self-similarity to prevent distortions.
    \item Comprehensive validation of our approach across two distinct setups: non-cascaded and cascaded video diffusion frameworks.
\end{itemize}
\vspace{-3pt}

\section{Background}
\label{background}

\subsubsection{Diffusion Models}
Diffusion models \cite{sohl2015deep, ho2020denoising, song2020score} define the generative process as the reverse of the forward noising process.
For clean data represented by $\boldsymbol{x}_0 \sim p_{\text{data}}(\boldsymbol{x})$, 
the forward process gradually introduces Gaussian noise through Markov transition with conditional densities 
\setlength{\abovedisplayskip}{3pt}
\setlength{\belowdisplayskip}{3pt}
\begin{equation}
\label{eq: forward}
\begin{split}
    p(\boldsymbol{x}_t \given \boldsymbol{x}_{t-1}) = 
    \mathcal{N} (\boldsymbol{x}_t \given \beta_t \boldsymbol{x}_{t-1}, (1-\beta_t) \boldsymbol{I}), \\
    p(\boldsymbol{x}_t \given \boldsymbol{x}_{0}) = \mathcal{N} (\boldsymbol{x}_t \given \sqrt{\bar \alpha} \boldsymbol{x}_0, (1-\bar \alpha) \boldsymbol{I}),
\end{split}
\end{equation}
where $\boldsymbol{x}_t \in \mathbb{R}^d$ is a noised latent representation at timestep $t$ and the noise schedule $\beta_t$ is a monotonically increasing sequence of $t$ with 
$\alpha_t := 1 - \beta_t$,  $\bar \alpha_t  := \prod_{i=1}^t \alpha_i$.
Then, the objective of diffusion model training is to obtain a multi-scale U-Net denoiser $\boldsymbol{\epsilon}_{\phi^*}$ that satisfies
\begin{equation}
    \label{eq: epsilon matching}
    \phi^* =  \arg \min_{\phi} \mathbb{E}_{\boldsymbol{x}_t \sim p_t(\boldsymbol{x}_t \given \boldsymbol{x}_0), \boldsymbol{x}_0 \sim p_{\text{data}}(\boldsymbol{x}_0), \boldsymbol{\epsilon} \sim \mathcal{N}(0, \boldsymbol{I})} \big[ \norm{\boldsymbol{\epsilon}_{\phi} (\boldsymbol{x}_t, t) - \boldsymbol{\epsilon}}_2^2 \big],
\end{equation}
where $\boldsymbol{\epsilon}_{\phi^*}(\boldsymbol{x}_t, t) \simeq\boldsymbol{\epsilon} = \frac{\boldsymbol{x}_t - \sqrt{\bar \alpha_t} \boldsymbol{x}_0 }{\sqrt{1-\bar \alpha}}$.
Notably, the Epsilon-Matching loss in \eqref{eq: epsilon matching} is equivalent to the Denoising Score Matching (DSM, \cite{hyvarinen2005estimation, vincent2011connection, song2019generative}) with alternative parameterization:
\begin{equation}
    \label{eq: score matching}
    \min_{\phi} \mathbb{E}_{\boldsymbol{x}_t, \boldsymbol{x}_0, \boldsymbol{\epsilon}} \big[ \norm{ \boldsymbol{s}_{\phi}^t (\boldsymbol{x}_t) - \nabla_{\boldsymbol{x}_t} \log p_t(\boldsymbol{x}_t \given \boldsymbol{x}_0) }_2^2 \big],
\end{equation}
where $\boldsymbol{s}_{\phi^*}(\boldsymbol{x}_t, t) \simeq - \frac{\boldsymbol{x}_t - \sqrt{\bar \alpha_t} \boldsymbol{x}_0 }{1-\bar \alpha} = - \frac{1}{\sqrt{1-\bar \alpha_t}} \boldsymbol{\epsilon}_{\phi^*} (\boldsymbol{x}_t, t)$.
For the reverse process, with the learned noise prediction network $\boldsymbol{\epsilon}_\phi^*$, the noisy sample of previous timestep $\boldsymbol{x}_{t-1}$ can be estimated by:
\begin{align}
    \label{eq: reverse sampling}
    \boldsymbol{x}_{t-1}=\frac{1}{\sqrt{\alpha_t}}\Big{(}\boldsymbol{x}_t-\frac{1-\alpha_t}{\sqrt{1-\bar{\alpha}_t}}\boldsymbol{\epsilon}_{\phi^*}(\boldsymbol{x}_t, t)\Big{)}+\tilde{\beta}_t \boldsymbol{\epsilon},
\end{align}
where $\tilde{\beta}_t := \frac{1 - \bar\alpha_{t-1}}{1 - \bar\alpha_t} \beta_t$ and $\boldsymbol{\epsilon} \sim\mathcal{N}(0,\boldsymbol{I})$.

\subsubsection{Conditional Generation}
In the context of conditional generation, data $\boldsymbol{x}$ is paired with an additional conditioning signal $y$, which in our case is a text caption.
To train a text-driven diffusion model, the text conditional embedding $y$ is incorporated into the objective as:
\begin{equation}
    \label{eq: condition epsilon matching}
    \min_{\phi} \mathbb{E}_{\boldsymbol{x}_t, \boldsymbol{x}_0, \boldsymbol{\epsilon}, y} \big[ \norm{\boldsymbol{\epsilon}_{\phi} (\boldsymbol{x}_t, t, y) - \boldsymbol{\epsilon}} \big]
\end{equation}
To augment the effect of text condition, classifier-free guidance \cite{ho2022classifier} attempts to benefit from both conditional and unconditional noise prediction, using a single network.
In specific, the epsilon prediction is defined as
\begin{equation}
\boldsymbol{\epsilon}_{\phi}^{w} (\boldsymbol{x}_{t}, t, y)
= (1+w) \boldsymbol{\epsilon}_{\phi} (\boldsymbol{x}_{t}, t, y)
- w \boldsymbol{\epsilon}_{\phi} (\boldsymbol{x}_{t}, t, \varnothing)
,
\label{eq:cfg}
\end{equation}
where $\varnothing$ denotes null text embedding and $w$ is the guidance scale.

\subsubsection{Video Diffusion Models}
Our framework leverages foundational video diffusion models for obtaining video scores.
Consider a video sequence of $N$ frames represented by $\boldsymbol{x}^{1:N} \in \mathbb{R}^{N \times d}$.
For any $n$-th frame within this sequence, denoted by $\boldsymbol{x}^{n} \in \mathbb{R}^{d}$, 
the noisy frame latent $\boldsymbol{x}_t^{n}$ sampled from $p_t(\boldsymbol{x}_t^n | \boldsymbol{x}^n)$ can be expressed as
$\boldsymbol{x}_t^{n} = \sqrt{\bar{\alpha}_t} \boldsymbol{x}^n + \sqrt{1 - \bar{\alpha}_t} \boldsymbol{\epsilon}_t^{n}$
, where $\boldsymbol{\epsilon}_t^n \sim \mathcal{N}(0, I)$.
Then, we similarly define $\boldsymbol{x}_t^{1:N}$ and $\boldsymbol{\epsilon}_t^{1:N}$.
The objective of video diffusion model training is then to obtain a denoiser network $\epsilon_{\phi^*}$ that satisfies:
\begin{equation}
    \label{eq: video epsilon matching}
    \phi^* =  \arg \min_{\phi} \mathbb{E}_{\boldsymbol{x}_t^{1:N}, \boldsymbol{x}^{1:N}, \boldsymbol{\epsilon}^{1:N}, y} \big[ \norm{\boldsymbol{\epsilon}_{\phi} (\boldsymbol{x}_t^{1:N}, t, y) - \boldsymbol{\epsilon}^{1:N}} \big],
\end{equation}
where $y$ is a text caption uniformly describing the video sequence $\boldsymbol{x}^{1:N}$.

Seeking to create videos that are both spatially and temporally enlarged and of high quality, video diffusion models have been expanded to cascaded pipelines \cite{ho2022video, ho2022imagen, singer2022make, blattmann2023align, wang2023lavie, zhang2023show}.
These cascaded video pipelines commonly follow a coarse-to-fine video generation approach, beginning with a module dedicated to creating keyframes that are low in both spatial and temporal resolution.
Subsequent stages involve temporal interpolation and spatial super-resolution modules, which work to increase the temporal and spatial resolution of the frames, respectively.
In this work, we plug our method into both cascaded and non-cascaded scenarios, proving its model-agnostic capability.

\section{DreamMotion}
\subsection{Overview}
\label{method: overview}
Starting with a series of input video frames $\boldsymbol{\hat{x}}^{1:N}$, a corresponding text prompt $\hat{y}$, and a target text $y$,
our goal is to get an edited video $\boldsymbol{x}^{1:N}$ that preserves the structural integrity and overall motion of $\boldsymbol{\hat{x}}^{1:N}$, while faithfully reflecting $y$.
DreamMotion starts by initializing the target video variable $\boldsymbol{x}_{0}^{1:N}(\theta)$ by the original video $\boldsymbol{\hat{x}}^{1:N}$.
Our optimization strategy is then three-pronged: 
(1) $\mathcal{L}_{\text{V-DDS}}$ that paints $\boldsymbol{x}_{0}^{1:N}(\theta)$ to match the appearance dictated by $y$, 
(2) $\mathcal{L}_{\text{S-SSM}}$ which encourages the structure of $\boldsymbol{x}_{0}^{1:N}(\theta)$ to align with $\boldsymbol{\hat{x}}^{1:N}$, 
(3) $\mathcal{L}_{\text{T-SSM}}$ which smoothens the gradients over the temporal dimension to eliminate any potential artifacts.

In Sec. \ref{method: inject}, we briefly review SDS and DDS loss formulations and describe how we directly modify the appearance of $\boldsymbol{x}^{1:N}$ with DDS-based gradients.
This technique, while effective in appearance injection, tends to accumulate structural inaccuracies, resulting in motion deviation in the end output. 
To address this, Sec. \ref{method: s-ssm} introduces a strategy for structural correction based on self-similarity, and Sec. \ref{method: t-ssm} details our approach for temporal smoothing, also leveraging self-similarity.
Finally, in Sec. \ref{method: cascaded}, we elaborate on the extension of DreamMotion to the cascaded video diffusion framework.
For simplicity, we primarily describe the diffusion model as operating in pixel space throughout this paper. However, in practice, our implementation encompasses both a latent space-based (Sec. \ref{sec: non-cascaded-comparison}, \cite{zeroscope}) and a pixel space-based video diffusion model (Sec. \ref{sec: cascaded-comparison}, \cite{zhang2023show}).

\begin{figure}[!t]
    \centering
    \includegraphics[width=\textwidth]
    {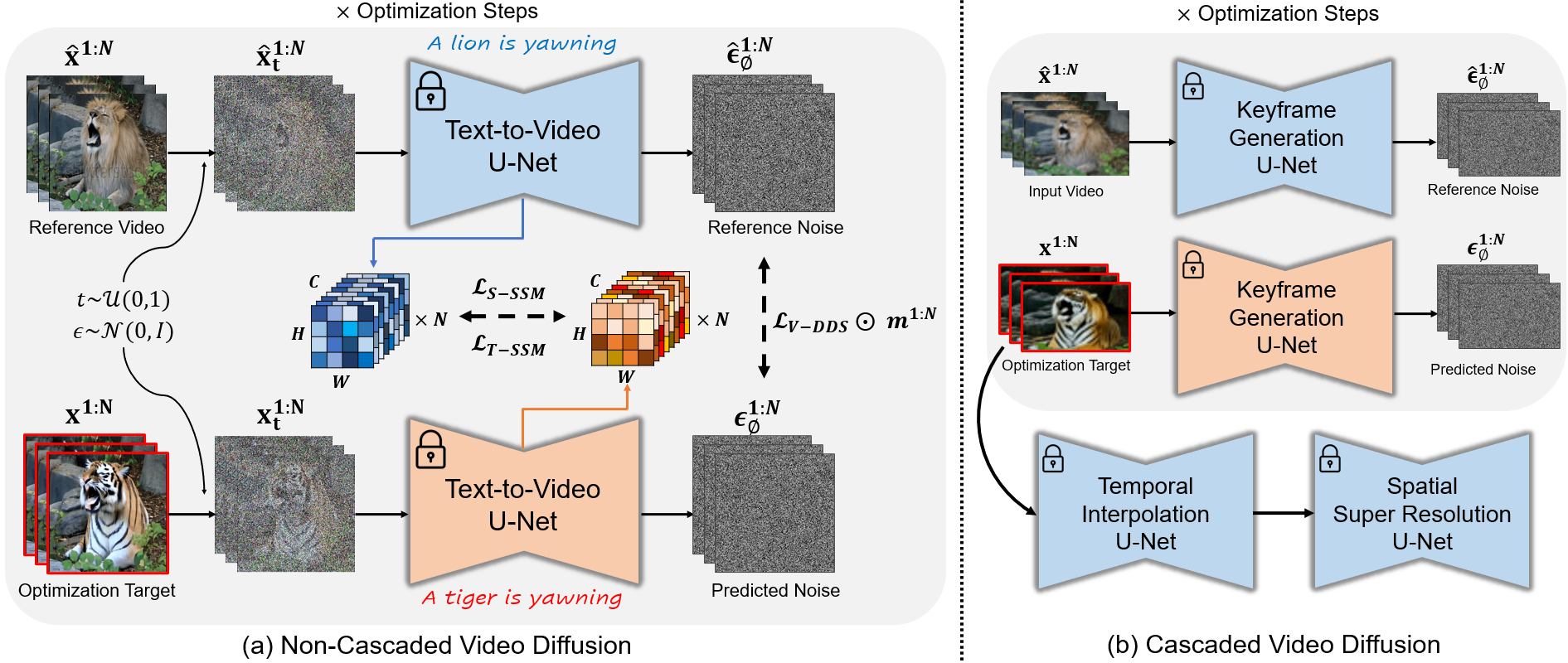}
    \caption{ 
    \textbf{Overview.} 
    DreamMotion leverages gradients derived from score distillation to inject target appearance, which is complemented by self-similarity alignments across spatial and temporal dimensions.
    This strategy seamlessly fits into cascaded video diffusion frameworks, confining the optimization on the keyframe generation phase.
    }
    \label{figure: overview}
\end{figure}

\subsection{Appearance Injection}
\label{method: inject}
\subsubsection{Image Score Distillation}
Let $\boldsymbol{x}_{0}(\theta)$ denote the target image parameterized by $\theta$ and $\boldsymbol{\epsilon}_{\phi}$ represent a T2I diffusion model.
SDS aims to align $\boldsymbol{x}_{0}(\theta)$ with the target text $y$ by optimizing the diffusion training loss gradient, expressed as:
\begin{equation}
    \label{eq: sds}
    \mathcal{L}_{\text{SDS}}(\theta; y) = \norm{\boldsymbol{\epsilon}_{\phi}^{w} (\boldsymbol{x}_{t}(\theta), t, y) - \boldsymbol{\epsilon}}_2^2,
\end{equation}
with $\epsilon \sim \mathcal{N}(0,\boldsymbol{I})$ and $t \sim \mathcal{U}(0,1)$.
Although $\nabla_{\theta} \mathcal{L}_{\text{SDS}}$ provides an efficient gradient term for incrementally refining the image fidelity to the text $y$, SDS often results in over-saturation, blurriness, and lack of details in the generated image \cite{wang2024prolificdreamer, lin2023magic3d, hertz2023delta, kim2023collaborative, nam2023contrastive}.

Under the assumption that the SDS score should be zero for pairs of correctly matched prompts and images, DDS \cite{hertz2023delta} enhances the gradient direction obtained from the SDS framework by incorporating an additional text-image pair, comprising a reference text $\boldsymbol{\hat{y}}$ and a reference image $\boldsymbol{\hat{x}}_0$.
Specifically, the noisy direction of the SDS score is calculated using the reference text-image branch, and this noisy score is subtracted from the main SDS optimization branch:
\begin{equation}
    \label{eq: dds}
    \mathcal{L}_{\text{DDS}}(\theta; y) = \norm{\boldsymbol{\epsilon}_{\phi}^{w} (\boldsymbol{x}_{t}(\theta), t, y) - \boldsymbol{\epsilon}_{\phi}^{w} (\boldsymbol{\hat{x}}_{t}, t, \hat{y})}_2^2.
\end{equation}

\vspace{-5pt}
\subsubsection{Video Score Distillation with Masked Gradients}
Leveraging a pre-trained T2V diffusion model $\boldsymbol{\epsilon}_{\phi}$, we extend the DDS mechanism to distill video scores.
Let $\boldsymbol{x}_{0}^{1:N}(\theta)$ represent the target video parameterized by $\theta$, and $\boldsymbol{x}_{0}^{1:N}$ denote the fixed, source video.
We optimize the video variable $\boldsymbol{x}_{0}^{1:N}(\theta)$ to reflect target text $y$ by minimizing:
\begin{equation}
    \label{eq: v-dds}
    \mathcal{L}_{\text{V-DDS}}(\theta; y) = \norm{\boldsymbol{\epsilon}_{\phi}^{w} (\boldsymbol{x}_{t}^{1:N}(\theta), t, y) - \boldsymbol{\epsilon}_{\phi}^{w} (\boldsymbol{\hat{x}}_{t}^{1:N}, t, \hat{y})}_2^2.
\end{equation}
While the video delta denoising score (V-DDS) offers a reliable gradient for gradually injecting appearance described by target text $y$, it still suffers from blurriness and over-saturation.
We mitigate this issue by additional mask conditioning.
Specifically, we filter the obtained gradients with a sequence of masks $m^{1:N}$ that annotate the objects to be edited in each frame, by $\nabla_{\theta} \mathcal{L}_{\text{V-DDS}} \odot m^{1:N}$.
The filtered gradients ensure that unintended regions in $\boldsymbol{x}_{0}^{1:N}(\theta)$ remain unaffected during V-DDS optimization (see Fig. \ref{figure: masking}).

A more significant issue arises when inaccurate gradients of $\mathcal{L}_{\text{V-DDS}}$ accumulate structural errors throughout the optimization process.
Unlike editing still images, these errors are particularly problematic in video editing,
as their accumulation deters temporal consistency within frames and often results in motion deflection, as illustrated in Fig. \ref{figure: progress}, \ref{figure: ablation_s_t}.
To tackle this, we propose to match self-similarities between target and reference branches, as detailed in Section \ref{method: s-ssm}.

\begin{figure}[!t]
    \centering
    \includegraphics[width=\textwidth]
    {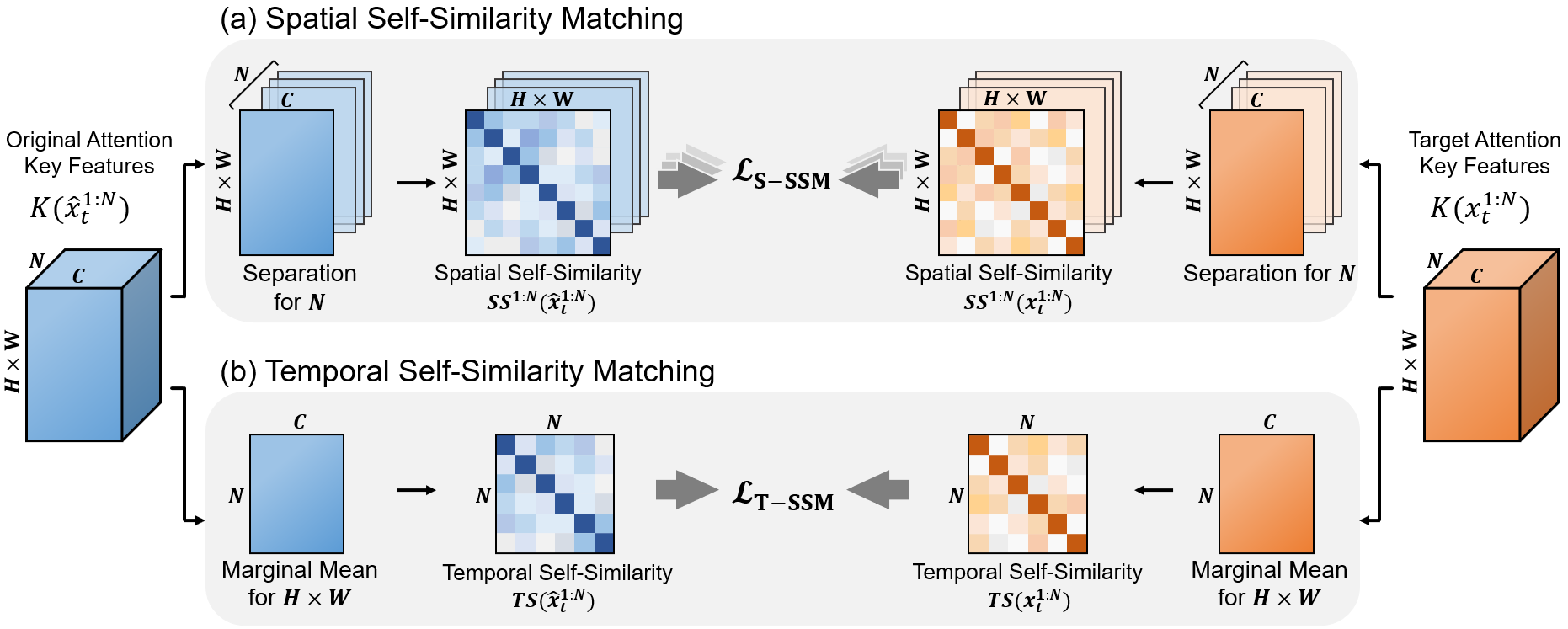}
    \caption{{
    \textbf{The proposed space-time self-similarity regularization:}
    (a) Spatial Self-Similarity Matching and (b) Temporal Self-Similarity Matching
    }}
    \label{figure: self-similarity}
\end{figure}

\subsection{Structure Correction}
\label{method: s-ssm}
\subsubsection{Spatial Self-Similarity Matching}
To address structural integrity, we require a representation that remains resilient against local texture patterns while retaining the global layout and overall shape of objects: self-similarity descriptors.
Self-similarity of visual features facilitates identifying objects by emphasizing the relationship of an object's appearance to its surroundings, rather than relying on its absolute appearance. 
This principle of relative appearance has been effectively applied across various domains: in traditional methods for matching visual patterns \cite{shechtman2007matching}, in the realm of neural style transfer through deep convolutional neural network features \cite{kolkin2019style}, and more recently, in the field of image editing utilizing DINO ViT features \cite{caron2021emerging, tumanyan2022splicing, kwon2022diffusion}.

Our contribution lies in pioneering the application of self-similarity through deep diffusion features \cite{tang2024emergent} to ensure structural correspondence between the target video $\boldsymbol{x}^{1:N}$ and the original video $\boldsymbol{\hat{x}}^{1:N}$.
To achieve this, we add \textit{identical noise} of timestep $t$ to both videos (Eq. \ref{eq: forward}), resulting in $\boldsymbol{x}_t^{1:N}$ and $\boldsymbol{\hat{x}}_t^{1:N}$, which are then feed-forwarded to the video diffusion U-Net $\boldsymbol{\epsilon}_{\phi}$ to extract a pair of attention key features $K(\boldsymbol{x}_t^{1:N}), K(\boldsymbol{\hat{x}}_t^{1:N}) \in \mathbb{R}^{N \times (H \times W) \times C} $.
Subsequently, we calculate spatial self-similarity map {$SS^n(\cdot) \in \mathbb{R}^{(H \times W) \times (H \times W)}$ of each $n$-th frame as follows:
\begin{equation}
\label{eq: ssm}
    SS^n_{i,j}(\boldsymbol{x}_t^{1:N}) = cos(K_i^n(x_t^{1:N}), K_j^n(x_t^{1:N})),
\end{equation}
}
where $cos(\cdot,\cdot)$ denotes the normalized cosine similarity, $i,j$ are all pairs of spatial indexes $(1 \leq i,j \leq (H {\times} W))$, and $\boldsymbol{x}_t^{1:N}(\theta)$ is simplified to $\boldsymbol{x}_t^{1:N}$ for brevity.
The spatial self-similarity matching objective is formulated as:
\begin{equation}
\label{eq: s-ssm}
    \mathcal{L}_{\text{S-SSM}}(\boldsymbol{x}^{1:N}_t, \boldsymbol{\hat{x}}^{1:N}_t) = 
    \frac{1}{N}
    \sum_{n=1}^{N} \norm{
    SS^{n}(\boldsymbol{x}_t^{1:N}) 
    - SS^{n}(\boldsymbol{\hat{x}}_t^{1:N}) 
    }_2^2,
\end{equation}
thereby quantifying and minimizing the discrepancy between the self-similarity maps of the target and original videos.

\begin{figure}[!t]
    \centering
    \includegraphics[width=\textwidth]
    {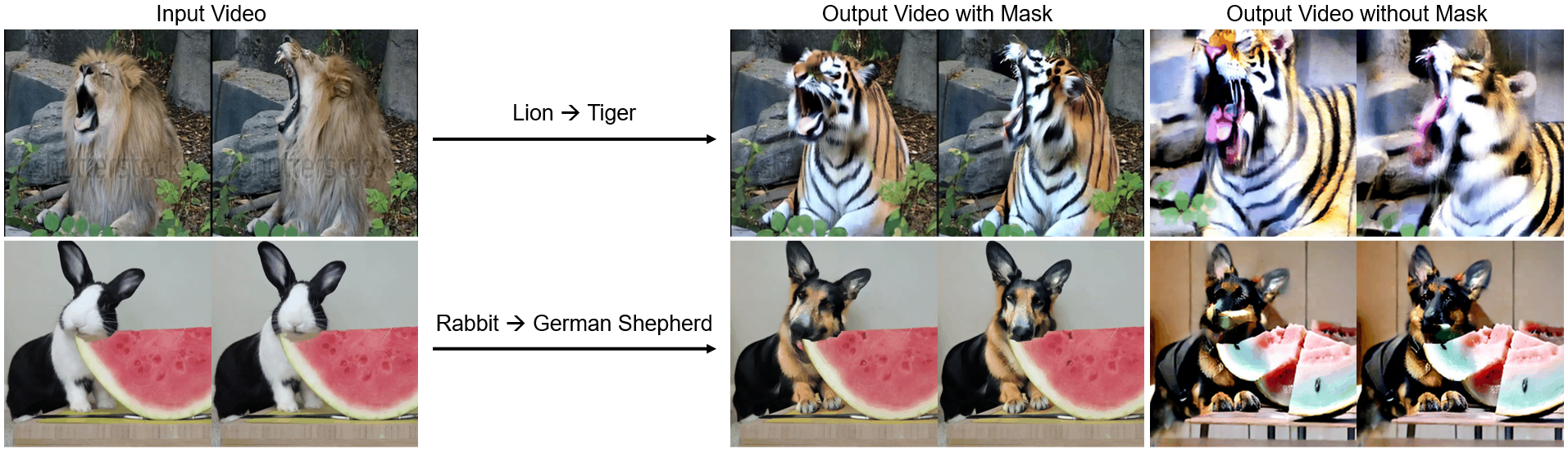}
    \caption{ 
    Filtering optimization gradients plays a crucial role in maintaining visual fidelity and preserving the structure of the input video. 
    Bounding boxes detected by off-the-shelf models \cite{li2022grounded, liu2023grounding} are used to create binary masks indicating the target regions for editing.
    }
    \label{figure: masking}
\end{figure}

\vspace{-3pt}
\subsection{Temporal Smoothing}
\label{method: t-ssm}
\vspace{-3pt}
\subsubsection{Temporal Self-Similarity Matching}
Although the spatial self-similarity alignment, facilitated by $\mathcal{L}_{\text{S-SSM}}$, proficiently maintains structural consistency between the original and modified videos, it operates as a frame-independent optimization method, without considering the temporal correlation between frames.
As observed in Fig. \ref{figure: ablation_s_t}, such per-frame operations can lead to localized distortions and notable flickering in the optimized frames.
To address these artifacts, we introduce a temporal regularization of $\mathcal{L}_{\text{S-SSM}}$ that models temporal correlations by leveraging self-similarity along the frame axis.

Calculating self-similarity over time necessitates a method to compress spatial information while retaining essential spatial details.
For this purpose, we employ spatial marginal mean a first-order statistic, spatial marginal mean, as our global descriptor. This choice is supported by prior works \cite{kolkin2019style, yatim2023space}, which have demonstrated their effectiveness in capturing crucial spatial details and serving as a robust global descriptor.
More concretely, we condense the spatial dimensions of the extracted key features 
$K(\boldsymbol{x}_t^{1:N}) \in \mathbb{R}^{N \times (H \times W) \times C}$ to $M[K(\boldsymbol{x}_t^{1:N})] \in \mathbb{R}^{N \times C}$ through the process defined as:
\begin{equation}
\label{eq: mean}
    M[K(\boldsymbol{x}_t^{1:N})] = 
    \frac{1}{H \cdot W}
    \sum_{i=1}^{H \cdot W} 
    K_{i}(\boldsymbol{x}_t^{1:N}),
\end{equation}
where $H$ and $W$ denote the height and width, respectively, and $C$ represents the channel dimension of the feature maps.
We then establish the temporal self-similarity $TS(\cdot) \in \mathbb{R}^{N \times N}$ as follows:
\begin{equation}
\label{eq: ssm-t}
    TS_{i,j}(\boldsymbol{x}_t^{1:N}) =
    cos(M_i[K(\boldsymbol{x}_t^{1:N})], M_j[K(\boldsymbol{x}_t^{1:N})]),
\end{equation}
where $i,j$ are from frame indexes $(1 \leq i,j \leq N)$.
Subsequently, the temporal self-similarity matching loss is formulated as:
\begin{equation}
\label{eq: t-ssm}
    \mathcal{L}_{\text{T-SSM}}(\boldsymbol{x}^{1:N}_t, \boldsymbol{\hat{x}}^{1:N}_t) = 
    \norm{
    TS(\boldsymbol{x}_t^{1:N}) 
    - TS(\boldsymbol{\hat{x}}_t^{1:N}) 
    }_2^2.
\end{equation}
It's noteworthy that the three losses $\mathcal{L}_\text{V-DDS}$, $\mathcal{L}_\text{S-SSM}$ and $\mathcal{L}_\text{T-SSM}$ share the same noise $\boldsymbol{\epsilon}$ and time $t$ for their computations, achieving a computationally efficient integration of optimizations through a single forward and reverse diffusion step.

\begin{figure}[!t]
    \centering
    \includegraphics[width=\textwidth]
    {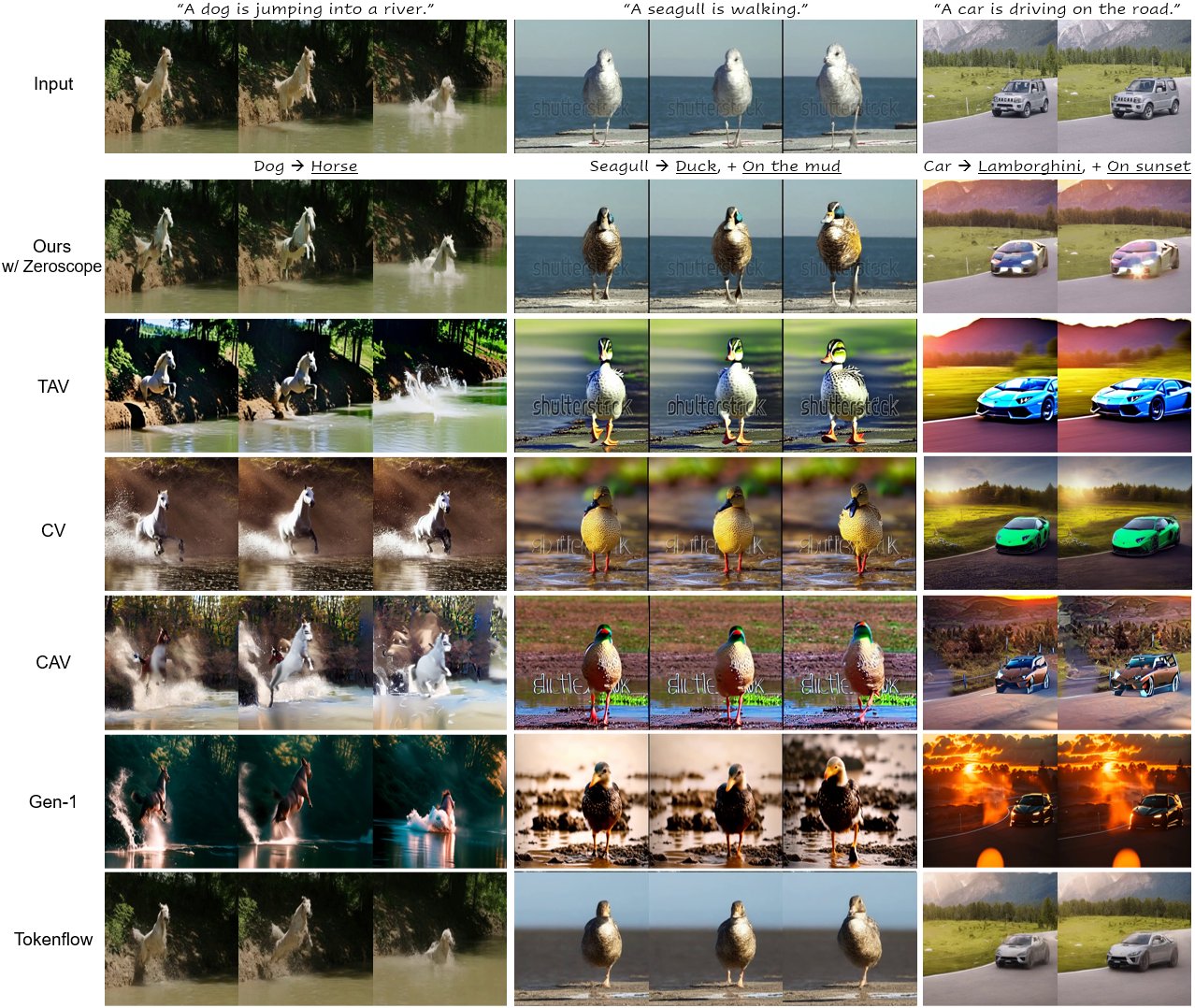}
    \caption{ 
    \textbf{Comparison.} 
    DreamMotion, applied to the Zeroscope model, is evaluated against five baseline methods. For a detailed assessment, please visit our project page.
    }
    \label{figure: comparison-non-cascaded}
\end{figure}

\subsection{Expansion to Cascade Video Diffusion}
\label{method: cascaded}
As outlined in Section \ref{background}, cascaded video diffusion models commonly utilize a coarse-to-fine approach for video generation, comprising three specialized modules that function in sequence: Keyframe Generation, Temporal Interpolation, and Spatial Super Resolution.
Rather than applying the optimization process through this comprehensive pipeline—a process that would result in prohibitively high computational costs—we focus our efforts exclusively on the initial Keyframe Generation stage. 
Within this approach,  we reinterpret $\boldsymbol{x}^{1:N}$ and $\boldsymbol{\hat{x}}^{1:N}$ to represent, respectively, the target and original keyframes, both resized to accommodate the low-resolution requirements of the keyframe generation space.
Furthermore, we designate $\boldsymbol{\epsilon}_{\phi}$ to represent the keyframe generation U-Net, excluding the temporal interpolation and super-resolution modules.
Following this setup, we apply our optimizations—$\mathcal{L}\text{V-DDS}$, $\mathcal{L}\text{S-SSM}$, and $\mathcal{L}\text{T-SSM}$—directly to $\boldsymbol{x}^{1:N}$.
After completing the optimization, these refined keyframes undergo further processing through the Temporal Interpolation and Spatial Super Resolution stages. This comprehensive procedure is depicted in Fig. \ref{figure: overview}-(b), illustrating the streamlined approach to integrating our optimization methods within the cascaded video diffusion model framework.

\section{Experiments}
\subsection{Non-cascaded Video Diffusion Framework}
\label{sec: non-cascaded-comparison}
\subsubsection{Setup}
For evaluation, we chose 26 text-video pairs from the public DAVIS \cite{pont20172017} and WebVid \cite{Bain21} datasets.
The videos vary in length from 8 frames to 16 frames.
In this experiment, we deploy our method on ZeroScope \cite{zeroscope}, a foundational text-to-video latent diffusion model.
The CFG scale $w$ is configured as 9.0.
We perform optimization for 200 steps using stochastic gradient descent (SGD) with a learning rate of 0.4.
The optimization of an 8-frame video requires approximately 2 minutes, while optimizing a 16-frame video takes around 4 minutes, utilizing a single A100 GPU.

\vspace{-4pt}
\subsubsection{Baselines}
Our method is evaluated alongside 1 one-shot and 4 zero-shot video editing baselines.
Tune-A-Video (TAV, \cite{wu2023tune}) selectively finetunes attention projection layers within an inflated T2I model on the given input video.
ControlVideo (CV, \cite{zhang2023controlvideo}) integrates temporally extended ControlNet \cite{zhang2023adding} to T2I diffusion and achieves motion-consistent video generation without any finetuning.
Both Control-A-Video (CAV, \cite{chen2023control}) and Gen-1 \cite{esser2023structure} are video diffusion models trained on large-scale text-image and text-video data.
They explicitly guide the ancestral denoising process with a series of structural conditions like depth maps.
Tokenflow \cite{geyer2023tokenflow} accomplishes time-consistent video editing by enforcing uniformity on the internal diffusion features across frames, in a zero-shot manner.

\vspace{-4pt}
\subsubsection{Qualitative Results}
Fig. \ref{figure: comparison-non-cascaded} offers a qualitative comparison between our method and state-of-the-art baselines; for complete videos, refer to our \textit{\href{https://hyeonho99.github.io/dreammotion}{project page}}.
Our method produces temporally consistent videos that closely adhere to the target prompt while most accurately preserving the motion of the input video, a feat that other baselines struggle to achieve simultaneously.

\vspace{-4pt}
\subsubsection{Quantitative Results}
We conducted a comprehensive quantitative evaluation, which includes both automatic metrics and a user study. The summarized results can be found in Tab. \ref{quantitative}.

\textit{(a) Automatic metrics.}
We first employ CLIP \cite{radford2021learning} to measure the text alignment and frame consistency of the edited videos. 
For assessing textual alignment \cite{hessel2021clipscore}, we measure average cosine similarity between the target text prompt and the edited frames. 
In terms of frame consistency, we calculate CLIP image features for every frame in the output video and then compute the average cosine similarity across all neighboring pairs of frames.
We additionally compute tracking-based motion fidelity score \cite{yatim2023space} and framewise LPIPS \cite{liu2024video} for measuring spatial consistency.
According to the results in Tab. \ref{quantitative}, our approach surpasses the baselines in achieving higher textual alignment and better spatial-temporal consistency.

\textit{(b) User study.} We surveyed 36 participants to assess the accuracy of editing, temporal consistency, and preservation of structure \& motion, using a rating scale from 1 to 5.
Participants were shown the input video followed by anonymized output videos from each baseline.
They were then asked the three questions:
{
(\lowercase\expandafter{\romannumeral1}) Edit Accuracy:
Does the output video accurately reflect the target text by appropriately editing all relevant elements?
}
{
(\lowercase\expandafter{\romannumeral2}) Frame Consistency:
Are the frames in the output video temporally consistent?
}
{
(\lowercase\expandafter{\romannumeral3}) Structure and Motion Preservation:
Has the structure and motion of the input video been accurately maintained in the output video?
}
Tab. \ref{quantitative} illustrates that our method outperforms the baselines in all measured aspects.
\vspace{-3pt}

\begin{table}
\centering
\resizebox{\linewidth}{!}{
\begin{tabular}{@{\extracolsep{0pt}}cccccccc@{}}
\toprule
& \multicolumn{4}{c}{Automatic Metrics} & \multicolumn{3}{c}{Human Evaluation} \\
\cmidrule(lr){2-5} \cmidrule(lr){6-8}
Method  & Text-Align \ & Frame-Con & Motion-Fidelity \ & Frame-LPIPS \ & Edit-Acc \ & Frame-Con \ & SM-Preserve \\
\cmidrule{1-8}
Tune-A-Video     & 0.8177 & 0.9218 & 0.6947 & 0.4172 & 3.52 & 2.82 & 2.89 \\
ControlVideo     & 0.7850 & 0.9678 &   -    & 0.3763 & 2.74 & 2.68 & 2.03 \\
Control-A-Video  & 0.7848 & 0.9297 & 0.8453 & 0.3829 & 2.17 & 2.16 & 2.18 \\
Gen-1            & \underline{0.8192} & \underline{0.9704} &   -    &    -   & 3.31 & \underline{3.62} & 2.95 \\
Tokenflow        & 0.7813 & 0.9576 & \underline{0.9184} & \underline{0.3427} & \underline{3.63} & 3.54 & \underline{3.92} \\
\rowcolor{Gray}
Ours (Zeroscope) & \textbf{0.8209} & \textbf{0.9726} & \bf 0.9259 & \bf 0.3042 & \textbf{4.14} & \textbf{4.21} & \textbf{4.33} \\
\bottomrule
\end{tabular}}
\caption{
\textbf{Quantitative evaluations.}
DreamMotion with Zeroscope outperforms various video editing methods in all seven features.
}
\label{quantitative}
\end{table}

\begin{figure}[!htb]
    \centering
    \includegraphics[width=\textwidth]
    {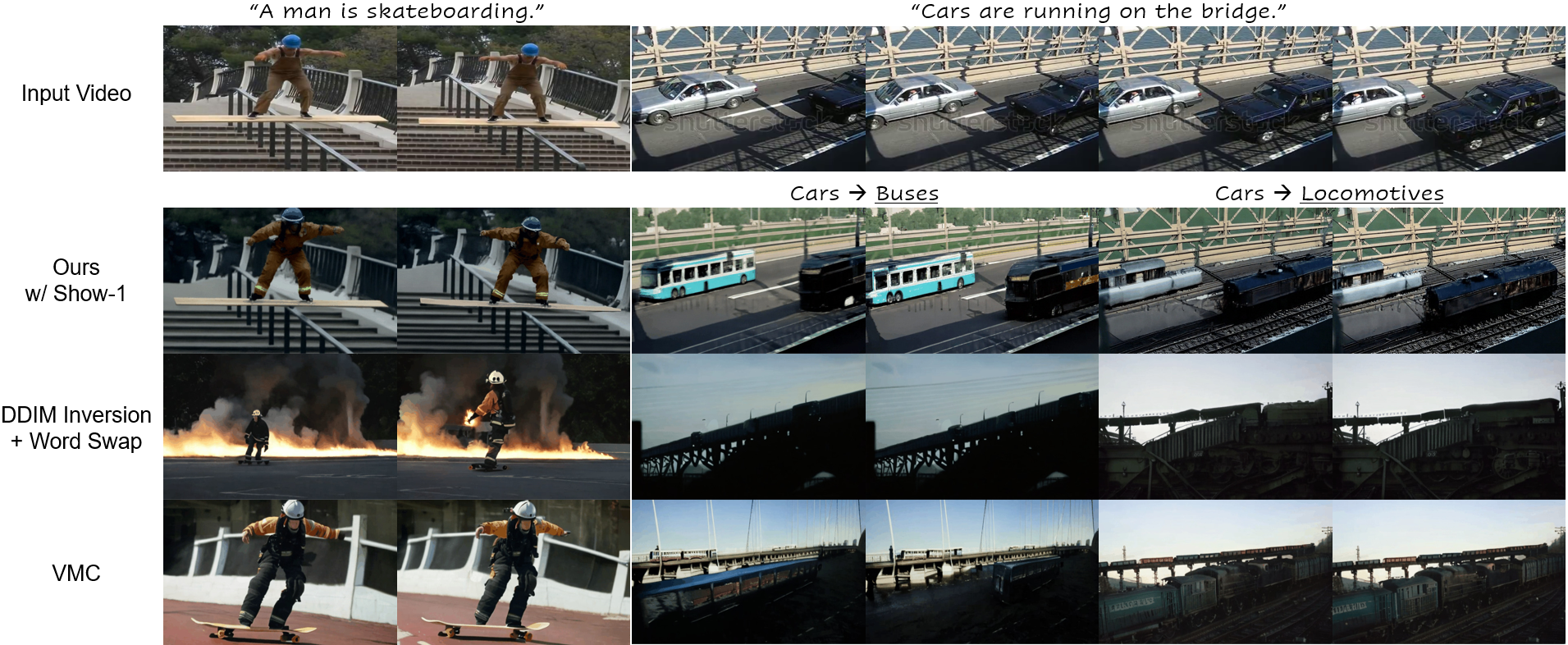}
    \caption{ 
    \textbf{Comparison.} 
    DreamMotion with Show-1 cascaded model is evaluated against two baselines.
    }
    \label{figure: comparison-cascaded}
\end{figure}

\begin{table}
\centering
\resizebox{\linewidth}{!}{
\begin{tabular}{@{\extracolsep{0pt}}cccccc@{}}
\toprule
& \multicolumn{2}{c}{Automatic Metrics} & \multicolumn{3}{c}{Human Evaluation} \\
\cmidrule(lr){2-3} \cmidrule(lr){4-6}
Method  & Text-Align \ & Frame-Con & Edit-Acc \ & Frame-Con \ & SM-Preserve \\
\cmidrule{1-6}
Inversion + Word Swap  & 0.7586 & 0.9714 & 3.36 & 3.42 & 2.21 \\
VMC              & 0.7563 & 0.9703 & 3.13 & 3.22 & 3.35 \\
\rowcolor{Gray}
Ours (Show-1)   & \textbf{0.7747} & \textbf{0.9755} & \textbf{3.97} & \textbf{3.74} & \textbf{4.30}  \\
\bottomrule
\end{tabular}}
\caption{
\textbf{Quantitative evaluations.}
DreamMotion utilizing Show-1 surpasses other cascaded baselines across the five features. Other baselines were also implemented using the same video model, ensuring a fair comparison.
}
\label{quantitative-cascaded}
\end{table}

\vspace{-6pt}
\subsection{Cascaded Video Diffusion Framework}
\label{sec: cascaded-comparison}
\subsubsection{Setup}
In this experiment, we utilize the 8-frame videos from the previously assembled text-video pairs.
Additionally, we benefit from Show-1 \cite{zhang2023show}, an open-source, cascaded video diffusion model.
As detailed in Sec. \ref{method: cascaded}, we compose our cascaded pipeline comprising Keyframe Generation, Temporal Interpolation, and Spatial Super Resolution, with all modules operating in pixel space.
Our method is implemented during the initial keyframe generation stage.
During keyframe optimization, these input videos undergo resizing to a resolution of 80x128 pixels, with the optimization process taking approximately 3 minutes on a single A100 GPU.
Following the optimization, the frame interpolation and super-resolution modules expand the output keyframes temporally and spatially, respectively.

\vspace{-5pt}
\subsubsection{Baselines}
To our knowledge, VMC \cite{jeong2024vmc} stands out as the sole video editing approach utilizing a cascaded video diffusion pipeline. 
VMC adapts temporal attention layers within the keyframe generation module, leveraging their novel motion distillation objective. 
For comparison purposes, we introduce an additional variant that employs direct inference using the cascaded pipeline with modified target text, starting from the DDIM inverted latents \cite{song2020denoising}.

\vspace{-5pt}
\subsubsection{Qualitative Results}
We qualitatively compare our method against baselines in Fig. \ref{figure: comparison-cascaded}. 
DreamMotion generates videos that match the structure and layout of the input video while adhering to the edit prompt, while other methods struggle to maintain the structural and motion integrity of the original video.
Since all three methods use unaltered temporal interpolation and super-resolution models after the generation of keyframes, they commonly produce temporally consistent videos.
For comprehensive results, please refer to the appendix.

\subsubsection{Quantitative Results}
Adopting the metrics outlined in Sec. \ref{sec: non-cascaded-comparison}, we compare our method quantitatively against baseline approaches, detailed in Tab. \ref{quantitative-cascaded}.
Notably, our approach demonstrated substantial superiority in Structure and Motion Preservation (SM-Preserve).

\begin{figure}[!t]
    \centering
    \includegraphics[width=\textwidth]
    {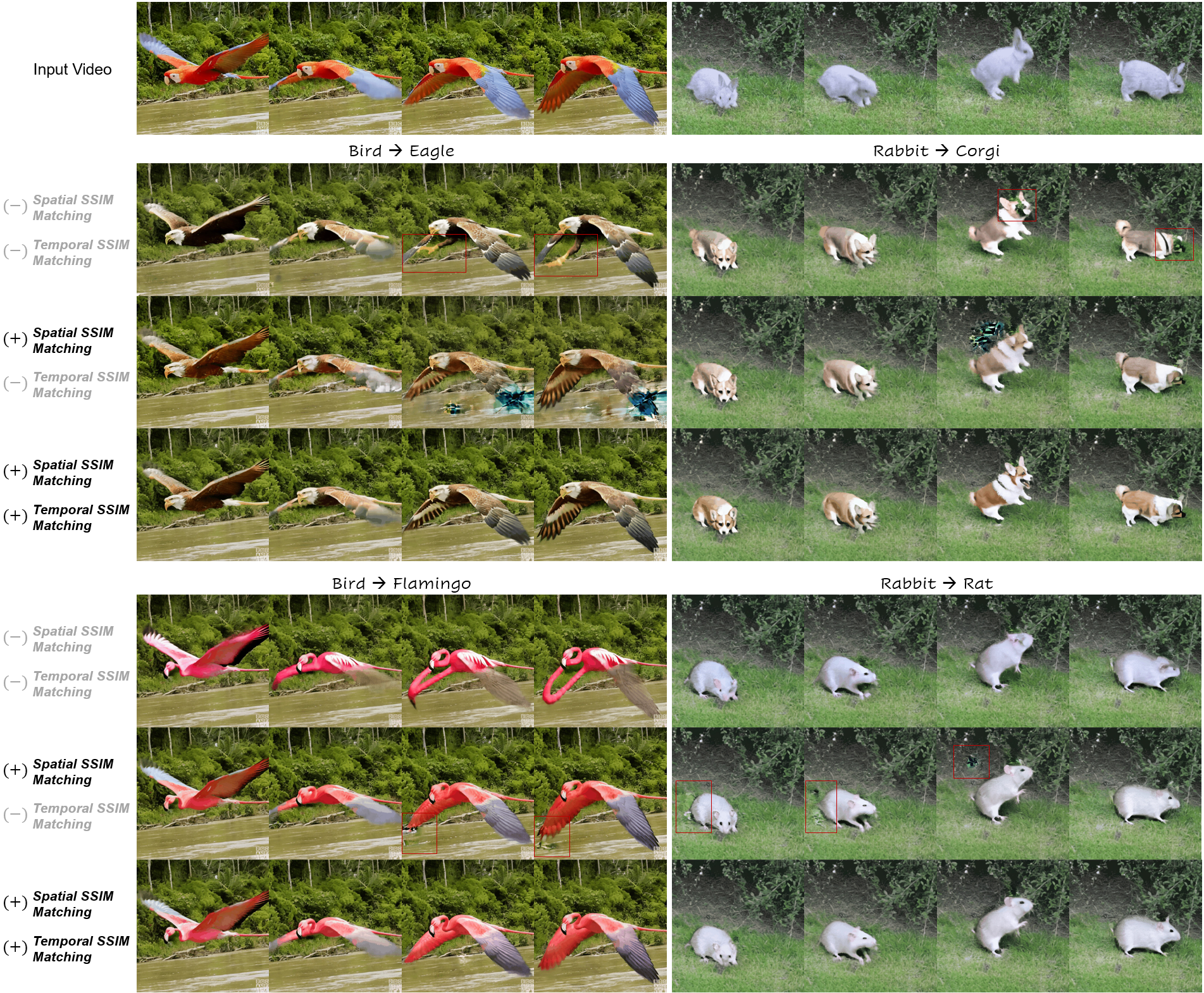}
    \caption{ 
    \textbf{Ablation of spatial and temporal self-similarity alignments.}
    Joint optimization of $\mathcal{L}_{\text{V-DDS}} + \mathcal{L}_{\text{S-SSM}} + \mathcal{L}_{\text{T-SSM}}$ generates the optimal output videos.}
    \label{figure: ablation_s_t}
\end{figure}

\begin{table}
\centering
\resizebox{\linewidth}{!}{
\begin{tabular}{@{\extracolsep{0pt}}ccccc@{}}
\toprule
      & Text-Align \ & Frame-Con & Motion-Fidelity \ & Frame-LPIPS\\
\cmidrule{1-5}
Ours \textit{wo $\mathcal{L}_\text{S-SSM}+ \mathcal{L}_\text{T-SSM}$} & 0.8202 & 0.9648 & 0.8426 & 0.3247 \\
Ours \textit{wo $\mathcal{L}_\text{T-SSM}$} & 0.8114 & 0.9567 & 0.9011 & 0.3186 \\
Ours \textit{wo masks} & 0.8180 & 0.9695  & 0.8653 & 0.3416 \\
Ours \textit{(full)} & \bf 0.8209 & \bf 0.9726 & \bf 0.9259 & \bf 0.3042 \\
\bottomrule
\end{tabular}}
\caption{
\textbf{Quantitative ablation.}
We demonstrate the impact of each factor by removing individual losses and masking conditions.
}
\label{quantitative-ablation}
\end{table}

\subsection{Ablation Studies}
In Fig. \ref{figure: masking}, we evaluate the impact of using bounding box-driven masks to selectively filter gradients during $\mathcal{L}_{\text{V-DDS}}$ update.
The results demonstrate that filtering gradients responsible for appearance injection enhances the precision of video editing and improves visual fidelity while avoiding issues of blurriness and saturation.

We next ablate the necessity of our self-similarity guidances.
Fig. \ref{figure: progress} illustrates the optimization progress with and without our self-similarity alignments.
The process begins with the initial input video (top row).
Solely using $\mathcal{L}_{\text{V-DDS}}$ for appearance injection (left) leads to the accumulation of structural errors as optimization progresses, resulting in motion deviation in the final output.
However, when the process is regularized by the spatial and temporal self-similarities (right), edited videos maintain the structure and motion fidelity throughout the optimization.
Additionally, in Fig. \ref{figure: ablation_s_t}, we illustrate video editing results under different optimization setups:
(\lowercase\expandafter{\romannumeral1}) $\mathcal{L}_{\text{V-DDS}}$.
(\lowercase\expandafter{\romannumeral2}) $\mathcal{L}_{\text{V-DDS}} + \mathcal{L}_{\text{S-SSM}}$.
(\lowercase\expandafter{\romannumeral2}) $\mathcal{L}_{\text{V-DDS}} + \mathcal{L}_{\text{S-SSM}} + \mathcal{L}_{\text{T-SSM}}$.
The absence of spatial self-similarity loss leads to inconsistency in object structures across frames.
For instance, the shape of a bird's wing varies, creating visible discrepancies, as shown in Fig. \ref{figure: ablation_s_t}-\textit{left}.
While aligning spatial self-similarity with the original video preserves structural integrity, it may generate artifacts in optimized areas. However, these artifacts are efficiently addressed through the addition of temporal self-similarity guidance. 
Lastly, Tab. \ref{quantitative-ablation} provides a quantitative analysis of each optimization term and masking condition.

\begin{figure}[t]
    \centering
    \includegraphics[width=\textwidth]
    {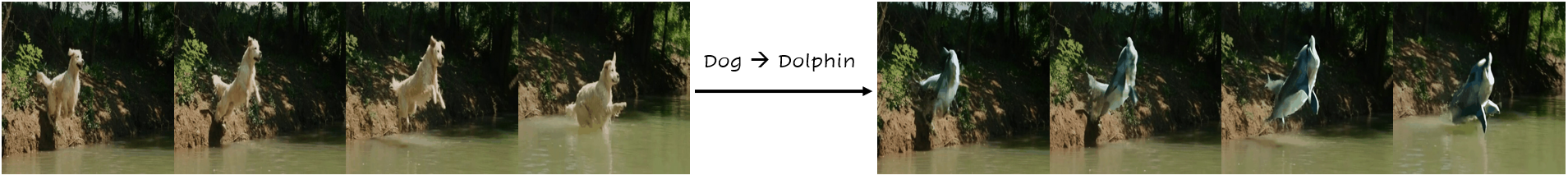}
    \caption{ 
    \textbf{Limitation.}
    DreamMotion limits its ability to produce videos that necessitate substantial structural alterations.
    }
    \label{figure: limitation}
\end{figure}

\vspace{-4pt}
\section{Conclusion}
\vspace{-3pt}
In this work, we have addressed the intricate challenge of diffusion-based video editing, a domain where formulating temporally consistent, real-world motion remains a notable obstacle. 
DreamMotion introduced score distillation-based optimization to text-to-video diffusion models, marking a departure from traditional, ancestral sampling-based video editing.
Our framework adeptly incorporated new content as specified by target text descriptions using the Video Delta Denoising Score, while preserving the the structural integrity and motion of the original video via a novel space-time self-similarity alignment.
Through rigorous validation in both cascaded and non-cascaded video diffusion settings, our approach has proven superior in maintaining the essence of the original video while seamlessly integrating desired alterations.
Regarding limitations, our framework is designed to preserve the structural integrity of the original video, and as such, it is not suited for edits that require significant structural changes (see Fig. \ref{figure: limitation}). 

\noindent\textbf{Ethics Statement}
Our work is based on generative models that carry the risk of being repurposed for unethical uses, such as misleading content.

\section*{Acknowledgments}
This work was supported by the National Research Foundation of Korea under Grant RS-2024-00336454.
This work was supported by Institute of Information \& communications Technology Planning \& Evaluation (IITP) grant funded by the Korea government (MSIT, Ministry of Science and ICT) (No. 2022-0-00984, Development of Artificial Intelligence Technology for Personalized Plug-and-Play Explanation and Verification of Explanation).
This work was supported by Institute of Information \& communications Technology Planning \& Evaluation (IITP) grant funded by the Korea government (MSIT) (No.2019-0-00075, Artificial Intelligence Graduate School Program (KAIST)).
This research was supported by Culture, Sports and Tourism R\&D Program through the Korea Creative Content Agency grant funded by the Ministry of Culture, Sports and Tourism in 2023.
This research was supported by Field-oriented Technology Development Project for Customs Administration funded by the Korea government (the Ministry of Science \& ICT and the Korea Customs Service) through the National Research Foundation (NRF) of Korea under Grant NRF2021M3I1A1097910.


%
%
\bibliographystyle{splncs04}
\bibliography{main}
\clearpage

\appendix
\section{Related Work}
\subsection{Video Editing using Diffusion Models}
Creating videos from textual descriptions necessitates ensuring realistic and temporally consistent motion, posing a unique set of challenges compared to text-driven image generative scenarios.
Before the advent of publicly accessible text-to-video diffusion models, Tune-A-Video \cite{wu2023tune} was at the forefront of one-shot-based video editing.
They proposed to inflate the image diffusion model to the pseudo video diffusion model by appending temporal modules to the image diffusion model \cite{rombach2022high} and reformulating spatial self-attention into spatio-temporal self-attention, facilitating inter-frame interactions.
However, the inflation often falls short of achieving consistent and complete motion, as motion preservation relies implicitly on the attention mechanism during inference.
Thus, the attention projection matrices within U-Net are often fine-tuned on the input videos \cite{wu2023tune, liu2024video, zhao2023controlvideo}. 
Utilizing explicit visual signals to steer the video denoising process is another common technique.
Pix2Video \cite{ceylan2023pix2video} and FateZero \cite{qi2023fatezero}, for instance, inject intermediate attention maps during the editing phase, which are derived during the input video inversion.
Others leverage pre-trained image adapter networks for structurally consistent video generation.
A notable example is ControlNet \cite{zhang2023adding}, which has been modified to accommodate a series of explicit structural indicators such as depth and edge maps.
Ground-A-Video \cite{jeongground} takes this a step further by adapting both ControlNet and GLIGEN \cite{li2023gligen} for video editing, utilizing spatially-continuous depth maps and spatially-discrete bounding boxes.

Despite the availability of open-source text-to-video (T2V) diffusion models \cite{wang2023modelscope, zeroscope, chen2023videocrafter1, wang2023lavie, zhang2023show}, recent endeavors frequently adopt a self-supervised strategy of fine-tuning pre-trained video generative models on an input video, to accurately capture intricate, real-world motion.
More specifically, several studies attempt to disentangle the appearance and motion elements of videos during the self-supervised fine-tuning.
For example, \cite{zhao2023motiondirector, wei2023dreamvideo, zhang2023motioncrafter} split the fine-tuning phase into two distinct pathways: one dedicated to integrating the subject's appearance into spatial modules, and the other aimed at embedding motion dynamics of a video into temporal modules within the T2V model.
Additionally, other studies \cite{jeong2024vmc, materzynska2023customizing, park2024spectral} attempt to extract and learn motion information from a single or a few reference videos.
VMC \cite{jeong2024vmc}, for instance, proposes to distill the motion within a video by calculating the residual vectors between consecutive frames, and refine temporal attention layers in cascaded video diffusion models.

Distinct from aforementioned approaches, DreamMotion circumvents the conventional ancestral sampling and employs Score Distillation Sampling \cite{poole2022dreamfusion} for editing appearance elements within a video.

\subsection{Visual Generation using Score Distillation Sampling}
Score Distillation Sampling (SDS) \cite{poole2022dreamfusion}, also known as Score Jacobian Chaining, has become the go-to method for text-to-3D generation in recent years \cite{wang2024prolificdreamer, metzer2023latent, lin2023magic3d, huang2023dreamtime, tsalicoglou2023textmesh, shen2021deep, chen2023fantasia3d, park2023ed}.
DreamFusion \cite{poole2022dreamfusion} first proposed to distill the generative prior of pre trained text-to-image models and optimize a parametric image synthesis model, such as NeRF \cite{mildenhall2021nerf}.
Despite its success, SDS often produces images that are overly saturated, blurry, and lack detail, largely due to the use of high CFG values \cite{wang2024prolificdreamer}.
To address these challenges, a range of derivative methods have been proposed \cite{wang2024prolificdreamer, metzer2023latent, lin2023magic3d, huang2023dreamtime, hertz2023delta, nam2023contrastive}.
Specifically, in the context of accurate image editing, DDS \cite{hertz2023delta} incorporates an additional reference branch with corresponding text to refine the noisy gradient of SDS.
Hifa \cite{zhu2023hifa}, instead, utilizes an estimated clean image rather than the predicted noise to compute denoising scores.
In our work, we employ a straightforward yet effective mask condition to refine DDS-generated gradients, allowing us to inject particular appearance into the video.
We further ensures the preservation of the video's original structure and motion through the novel regularization of space-time self-similarity alignment.

\section{Technical Details}
For the sampling of timestep $t$ to derive $\boldsymbol{x}_t^{1:N}$ and $\boldsymbol{\hat{x}}_t^{1:N}$, we restrict $t$ to the range $t \sim \mathcal{U}(0.05, 0.95)$, in line with DDS's official implementation\footnote{\url{https://github.com/google/prompt-to-prompt/blob/main/DDS_zeroshot.ipynb}}.
For the extraction of attention key features from video diffusion U-Net, we specifically select the self-attention layers within its decoder part.
In the non-cascaded video diffusion experiments, we utilize Zeroscope\footnote{\url{https://huggingface.co/cerspense/zeroscope_v2_576w}} \cite{zeroscope}, a diffusion model that operates in latent-space rather than pixel-space. 
Practically, this means the video frames are initially encoded into latent representations by VAEs, and then our proposed optimizations take place within this latent space.
Conversely, in experiments involving the cascaded video diffusion framework, we select Show-1\footnote{\url{https://huggingface.co/showlab/show-1-base}} \cite{zhang2023show}, where the keyframe generation UNet of Show-1 uses a pixel-space diffusion. 
As a result, the video frames stay in pixel space, with optimizations carried out directly within this domain.

To produce output videos using Tune-A-Video \cite{wu2023tune}, ControlVideo \cite{zhang2023controlvideo}, Control-A-Video \cite{chen2023control}, and TokenFlow \cite{geyer2023tokenflow}, we utilized the official github repositories along with their default hyperparameters.
The results from Gen-1 \cite{esser2023structure} were generated using their web-based product.
Given that Gen-1 generates videos with temporally extended sequences, including duplicated frames, we removed these repeated frames when calculating CLIP-based frame consistency to ensure a fair evaluation.
However, for the human evaluation, the outputs from Gen-1 were used as is, without any modifications.

\begin{figure}[t]
    \centering
    \includegraphics[width=\textwidth]
    {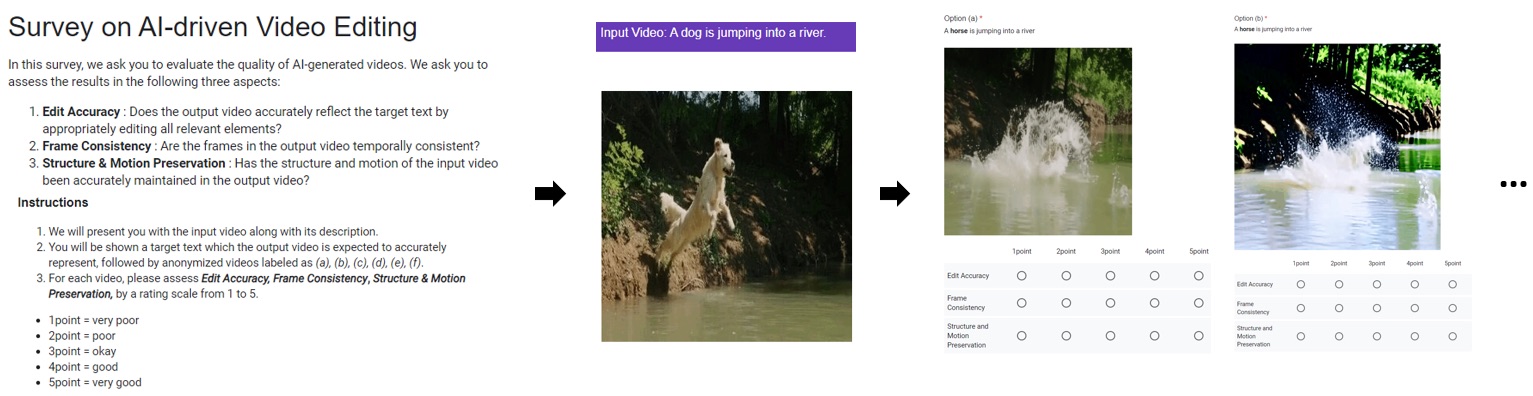}
    \caption{ 
    Interface of human evaluation.
    }
    \label{supple: human}
\end{figure}

\section{User Study Interface}
We carried out human evaluations to assess various methods based on three key aspects: Edit Accuracy, Frame Consistency, and Sturcture \& Motion Preservation.
Initially, we present the input video alongside its text description, as shown in Figure \ref{supple: human}.
Subsequently, we display target text with anonymized videos generated by each method and ask participants to evaluate them across the aforementioned three criteria.
The human evaluation results, detailed in Table 1 of the manuscript, unequivocally highlight the superiority of DreamMotion in both video diffusion frameworks.

\begin{figure}[t]
    \centering
    \includegraphics[width=\textwidth]
    {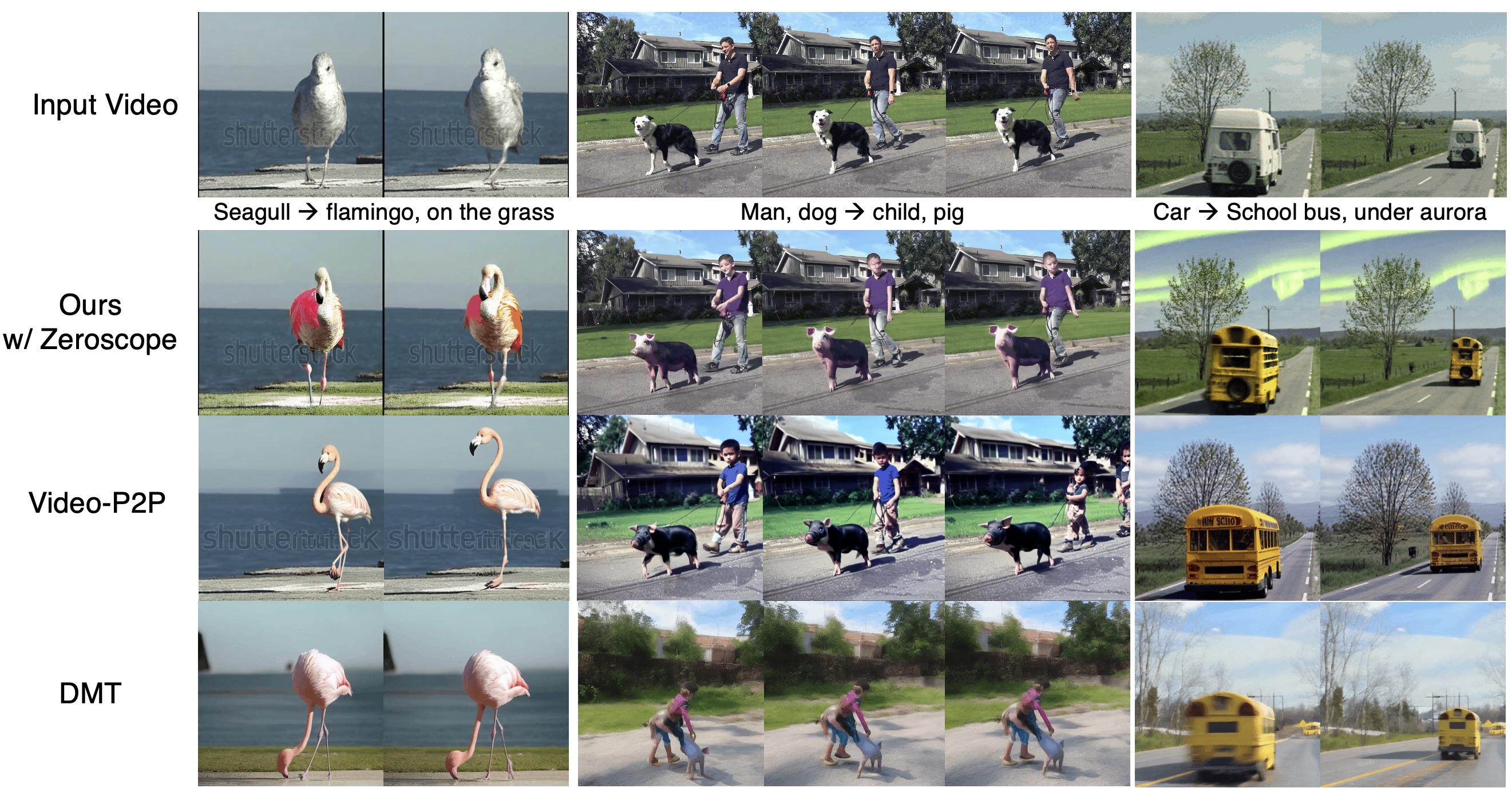}
    \caption{ 
    Additional qualitative comparison with DMT and Video-P2P.
    }
    \label{supple: qualitative-comparison}
\end{figure}

\begin{wraptable}{r}{0.5\textwidth}
\centering
\resizebox{6cm}{!}{
\begin{tabular}{@{\extracolsep{0pt}} c|ccc}
\hline
                & Video-P2P & DMT & Ours (Zeroscope) \\
\hline
Motion-Fidelity & 0.7384 & 0.8697 & \bf 0.9259 \\
Frame-LPIPS     & 0.3395 & 0.3078 & \bf 0.3042 \\
\hline
\end{tabular}
}
\caption{
Additional quantitative comparison with DMT and Video-P2P.
}
\label{supple: quantitative-comparison}
\end{wraptable}

\section{Additional Comparison}
We additionally compared our method with two video editing techniques specifically designed for localized editing: Video-P2P \cite{liu2024video} and Diffusion-Motion-Transfer (DMT) \cite{yatim2023space}. For qualitative comparison, see Fig. \ref{supple: qualitative-comparison}. For quantitative comparison in Tab. \ref{supple: quantitative-comparison}, we employed tracking-based motion fidelity score \cite{yatim2023space} and framewise LPIPS \cite{liu2024video} to evaluate spatial consistency.

\section{Additional Results}
\begin{figure}[htb]
    \centering
    \includegraphics[width=\textwidth]
    {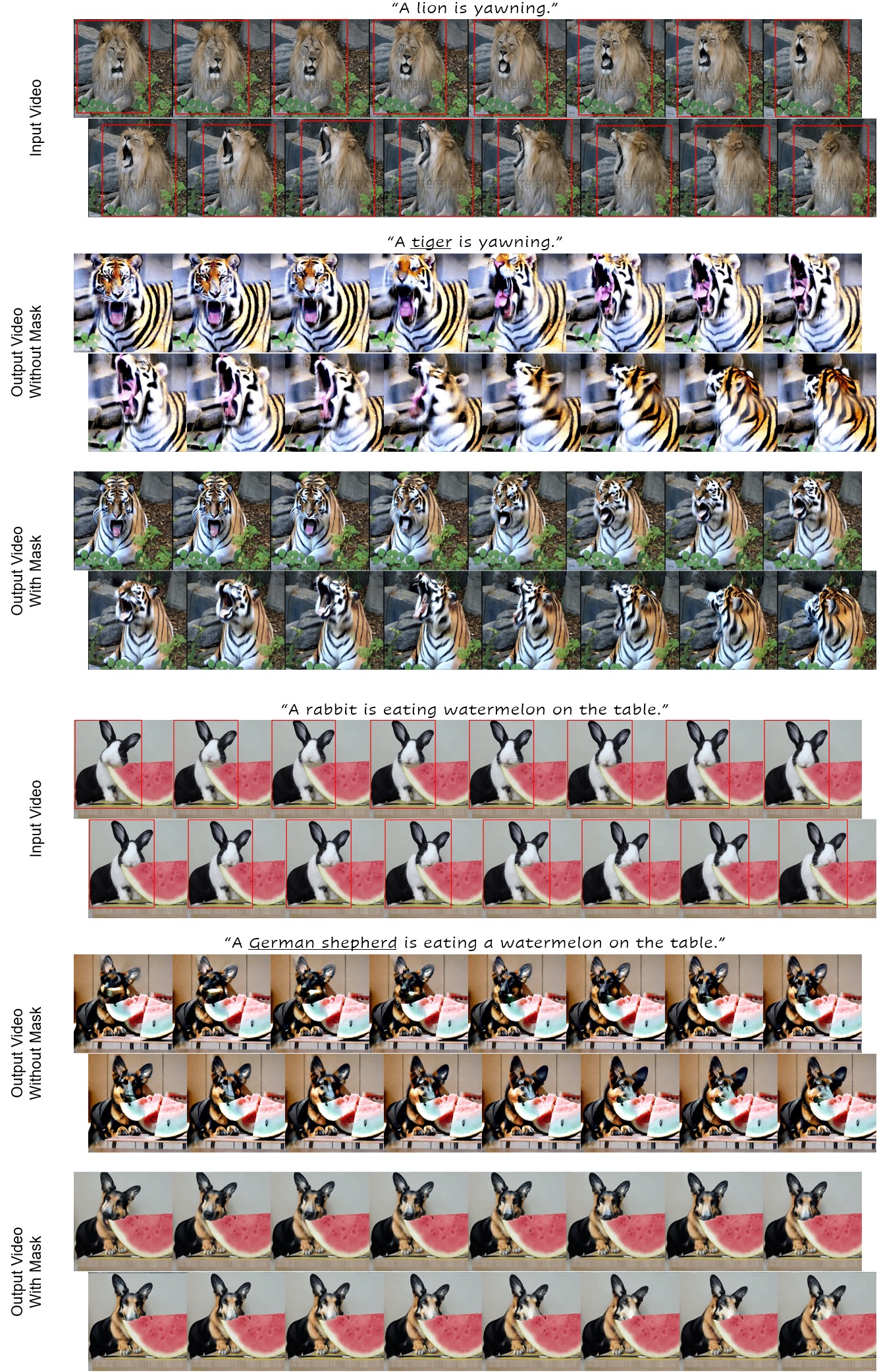}
    \caption{ 
    Video optimization with and without masking gradients.
    }
    \label{supple: mask}
\end{figure}

\begin{figure}[htb]
    \centering
    \includegraphics[width=\textwidth]
    {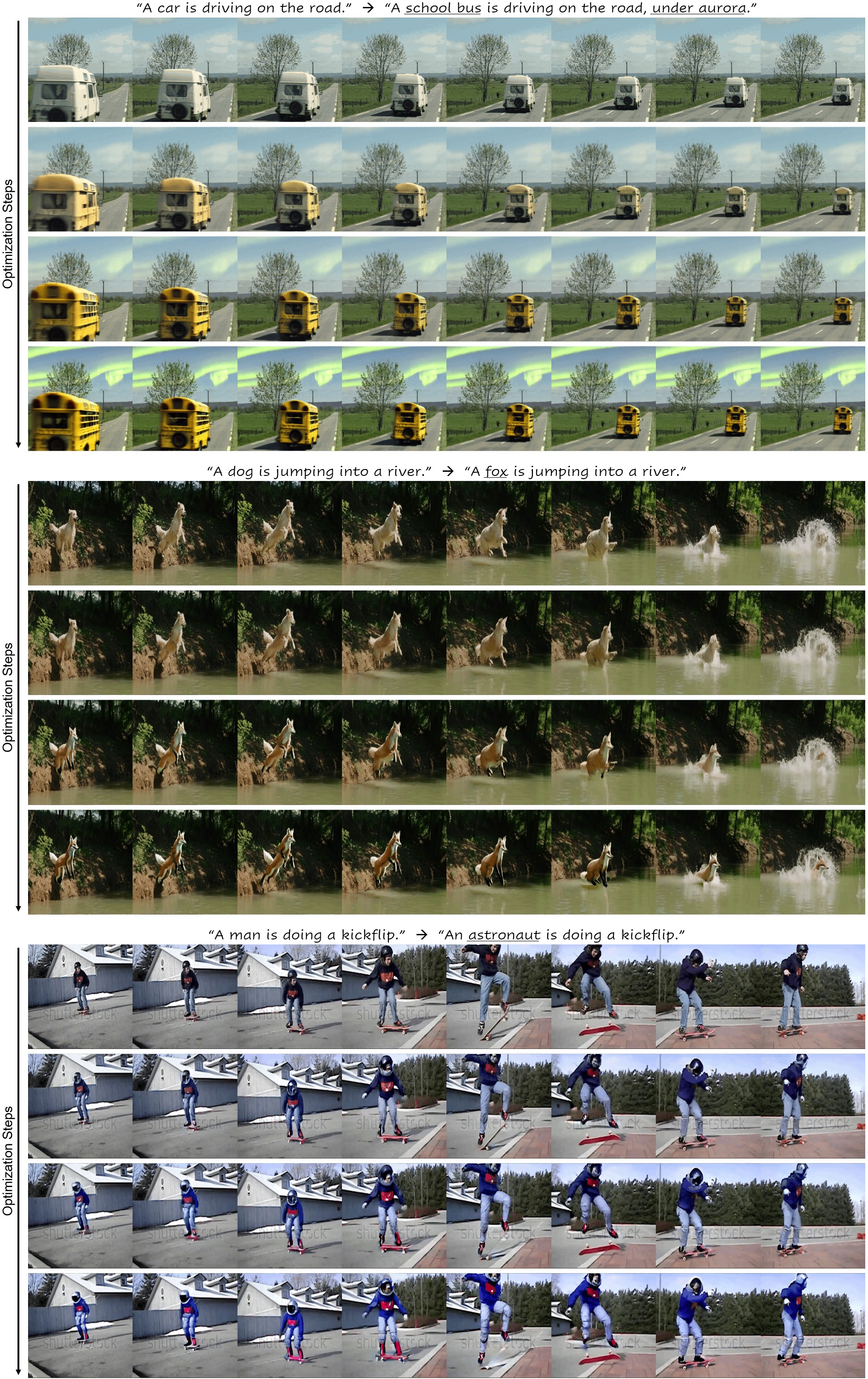}
    \caption{ 
    Visualization of optimization progress.
    }
    \label{supple: progress}
\end{figure}

\begin{figure}[htb]
    \centering
    \includegraphics[width=\textwidth]
    {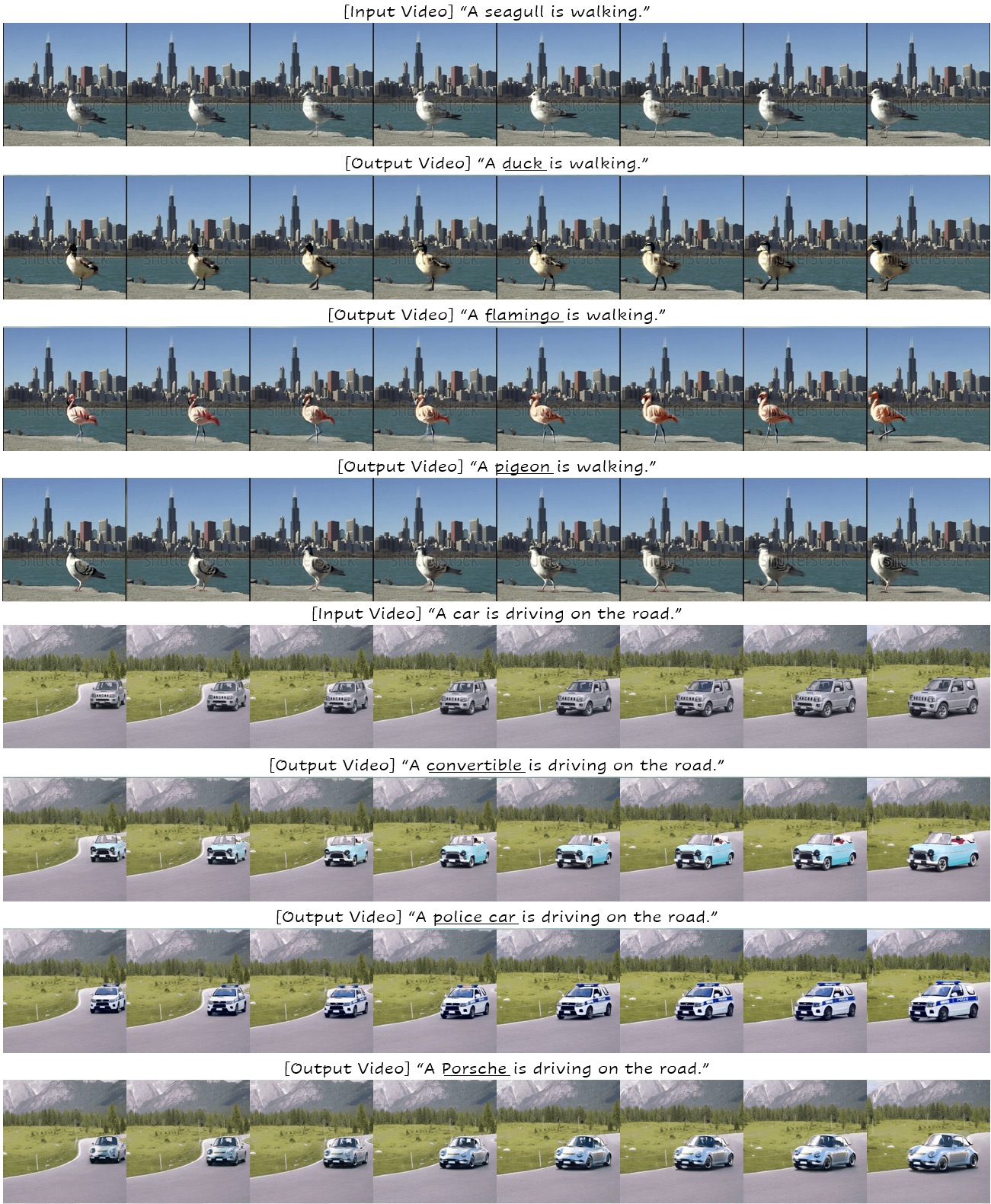}
    \caption{ 
    Additional results of DreamMotion with Zeroscope T2V.
    }
    \label{supple: zero1}
\end{figure}

\begin{figure}[htb]
    \centering
    \includegraphics[width=\textwidth]
    {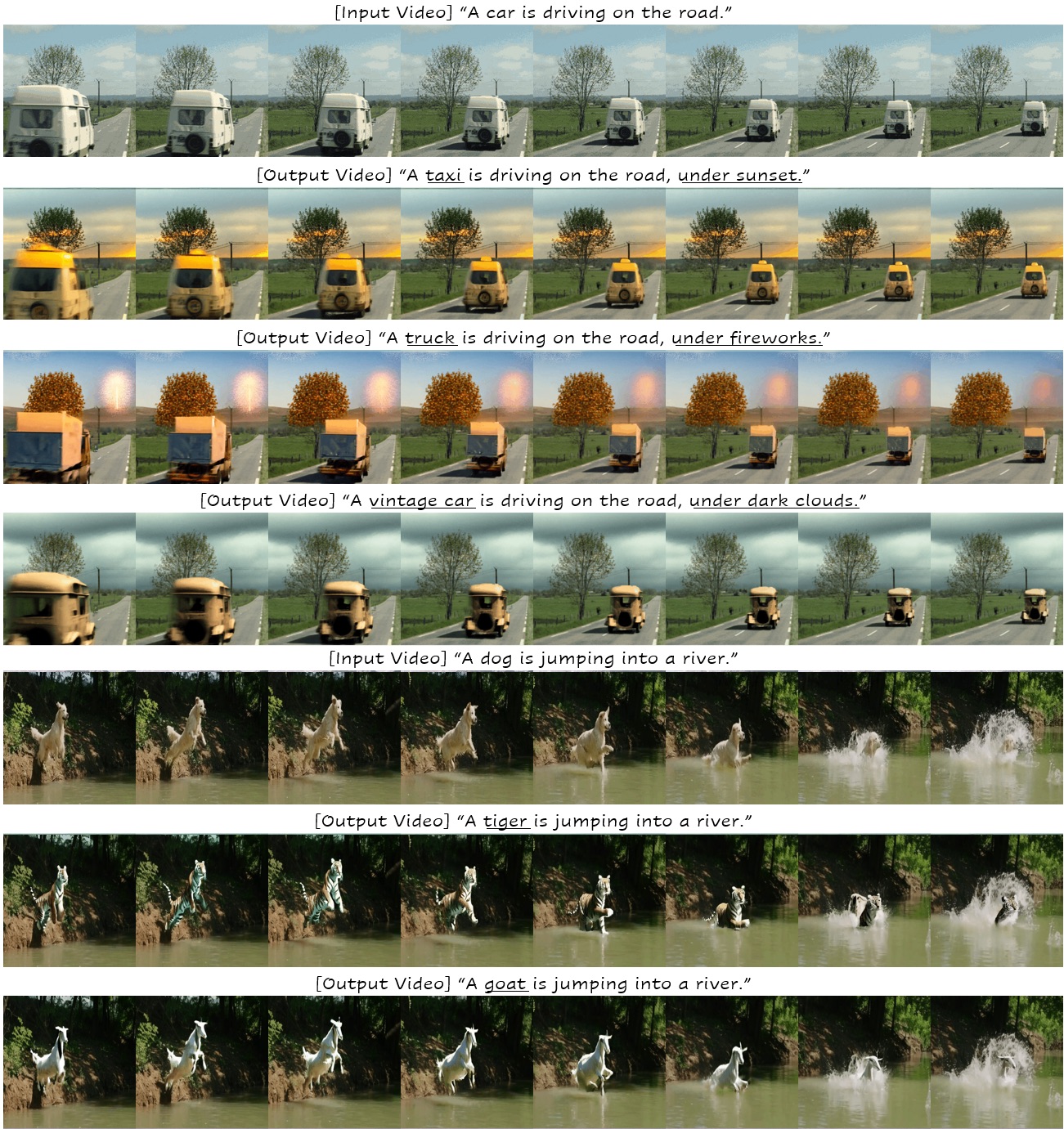}
    \caption{ 
    Additional results of DreamMotion with Zeroscope T2V.
    }
    \label{supple: zero2}
\end{figure}

\begin{figure}[htb]
    \centering
    \includegraphics[width=\textwidth]
    {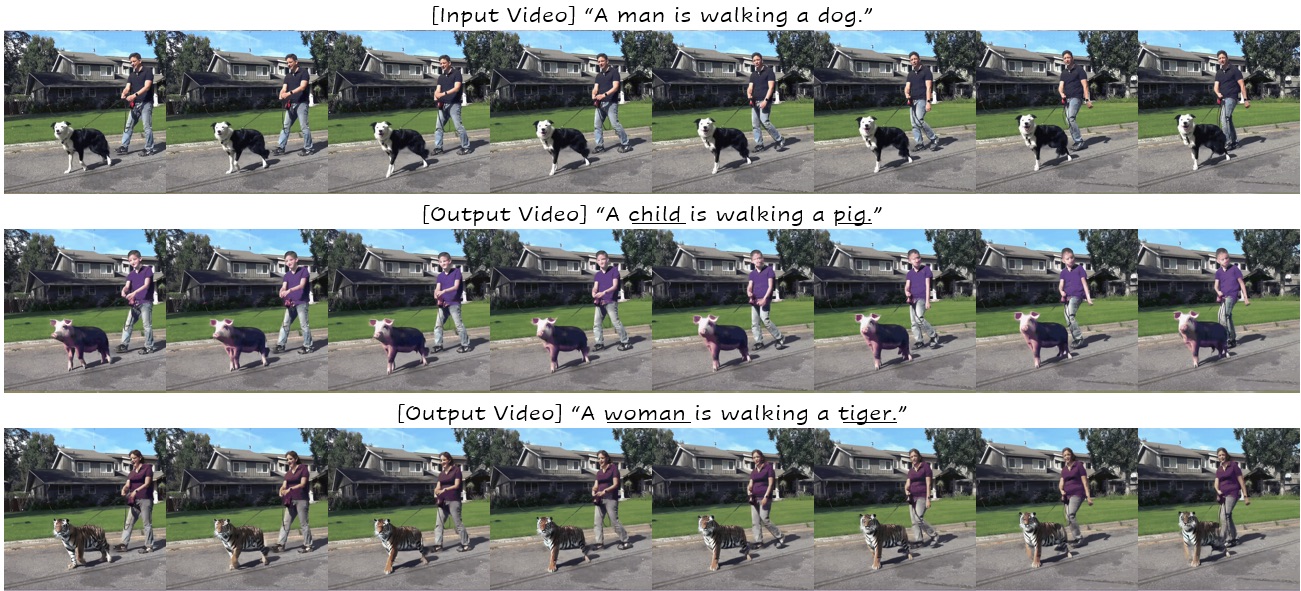}
    \caption{ 
    Additional results of DreamMotion with Zeroscope T2V.
    }
    \label{supple: zero3}
\end{figure}

\begin{figure}[htb]
    \centering
    \includegraphics[width=\textwidth]
    {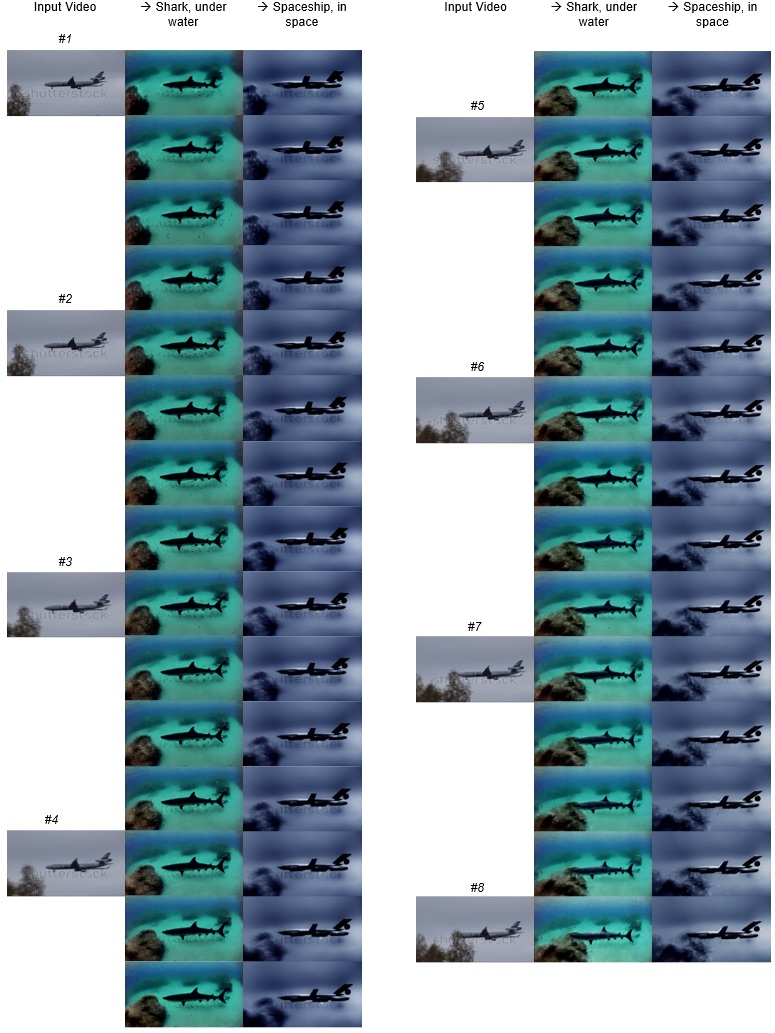}
    \caption{ 
    Additional results of DreamMotion with Show-1 Cascade.
    }
    \label{supple: show1}
\end{figure}

\begin{figure}[htb]
    \centering
    \includegraphics[width=\textwidth]
    {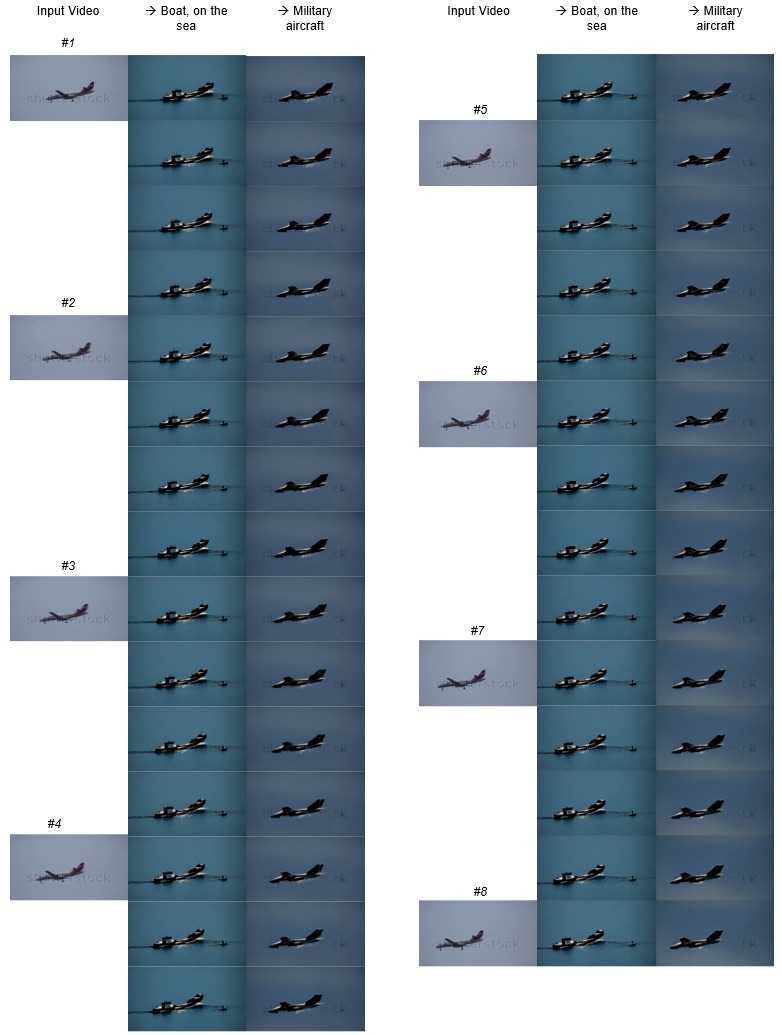}
    \caption{ 
    Additional results of DreamMotion with Show-1 Cascade.
    }
    \label{supple: show2}
\end{figure}

\begin{figure}[htb]
    \centering
    \includegraphics[width=\textwidth]
    {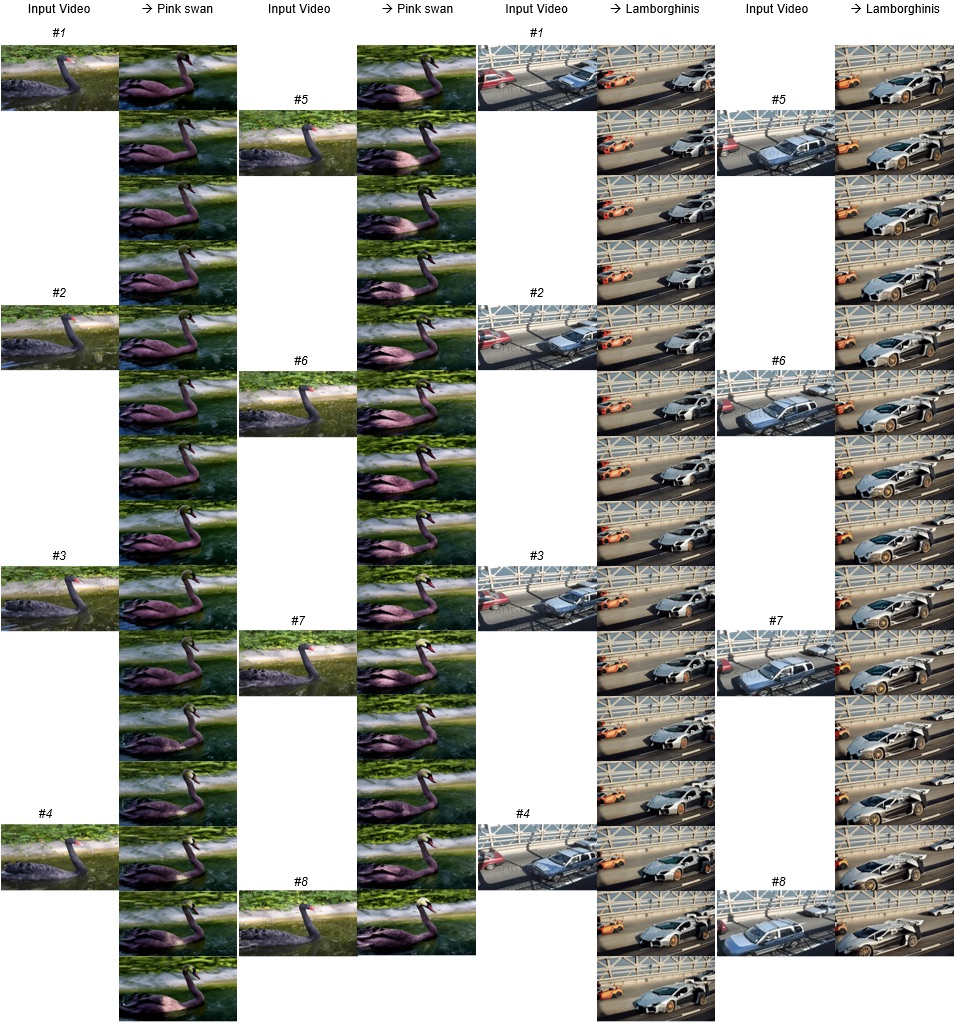}
    \caption{ 
    Additional results of DreamMotion with Show-1 Cascade.
    }
    \label{supple: show3}
\end{figure}

This section is dedicated to presenting additional outcomes of DreamMotion.
Figure \ref{supple: mask} offers a comprehensive view of the results from Figure 6 in the main paper, demonstrating the effect of masking DDS-driven gradients.
Annotations within the input video frames indicate the masks used.
In Figure \ref{supple: progress}, we present the progress of DreamMotion optimization by visualizing intermediate output videos.
Figures \ref{supple: zero1}, \ref{supple: zero2}, and \ref{supple: zero3} showcase input and corresponding edited videos generated with DreamMotion on Zeroscope, using various target prompts.
To accommodate space constraints, only odd or even frames from 16-frame videos are selected for display.
Figures \ref{supple: show1}, \ref{supple: show2}, and \ref{supple: show3} feature videos edited by DreamMotion on the Show-1 Cascade model \cite{zhang2023show}, with the left columns displaying 8-frame input videos and the adjacent columns showing 29-frame output videos.
Our qualitative results are uploaded on our \href{https://hyeonho99.github.io/dreammotion/}{project page}.

\end{document}